\documentclass[preprint,5p]{elsarticle}
\usepackage{graphicx} 
\usepackage[utf8]{inputenc}
\usepackage{amsmath}
\usepackage{enumitem}
\usepackage{geometry}
\usepackage{microtype}

\usepackage[most]{tcolorbox}
\usepackage{listings}
\usepackage{bera} 
\usepackage{colortbl}

\usepackage{tikz}
\usetikzlibrary{shapes, arrows.meta, positioning, shadows, calc}
\usetikzlibrary{shapes, arrows.meta, positioning, shadows, decorations.pathmorphing}
\usepackage{xcolor}
\usetikzlibrary{shapes.geometric, arrows.meta, positioning, fit, decorations.pathreplacing}

\definecolor{iotblue}{RGB}{52, 152, 219}
\definecolor{llmgreen}{RGB}{46, 204, 113}
\definecolor{ragpurple}{RGB}{155, 89, 182}
\definecolor{standardorange}{RGB}{230, 126, 34}
\definecolor{conflictred}{RGB}{231, 76, 60}
\definecolor{lightgray}{RGB}{236, 240, 241}

\definecolor{delim}{RGB}{20,105,176}
\definecolor{numb}{RGB}{106, 109, 32}
\definecolor{string}{rgb}{0.64,0.08,0.08}

\lstdefinelanguage{json}{
    frame=single,
    rulecolor=\color{black},
    showspaces=false,
    showtabs=false,
    breaklines=true,
    upquote=true,
    morestring=[b]",
    stringstyle=\color{string},
    literate=
     *{0}{{{\color{numb}0}}}{1}
      {1}{{{\color{numb}1}}}{1}
      {2}{{{\color{numb}2}}}{1}
      {3}{{{\color{numb}3}}}{1}
      {4}{{{\color{numb}4}}}{1}
      {5}{{{\color{numb}5}}}{1}
      {6}{{{\color{numb}6}}}{1}
      {7}{{{\color{numb}7}}}{1}
      {8}{{{\color{numb}8}}}{1}
      {9}{{{\color{numb}9}}}{1}
      {\{}{{{\color{delim}{\{}}}}{1}
      {\}}{{{\color{delim}{\}}}}}{1}
      {[}{{{\color{delim}{[}}}}{1}
      {]}{{{\color{delim}{]}}}}{1},
}
\makeatletter
\def\lst@lettertrue{\let\lst@ifletter\iffalse}
\makeatother

\usepackage{amssymb}
\usepackage{booktabs}
\usepackage{multirow, makecell}
\usepackage{multicol}

\newcommand{\inlinecode}[1]{%
  \texttt{%
    \setlength{\spaceskip}{0.5em plus 1em minus 0.1em}%
    \ifdim\lastskip>0pt \unskip\hspace{0.5em plus 0.5em minus 0.1em}\fi
    #1
  }
}

\usepackage[acronym]{glossaries}

\makeglossaries

\newacronym{rs}{RS}{reasoning-specialized}
\newacronym{cs}{CS}{code-specialized}
\newacronym{dm}{DM}{data model}
\newacronym{sdm}{SDM}{smart data model}
\newacronym{gp}{GP}{general purpose}
\newacronym{llm}{LLM}{large language model}
\newacronym{er}{ER}{entity resolution}
\newacronym{sem}{SEM}{Subject Exact Match}
\newacronym{dmem}{DMEM}{Data Model Exact Match}
\newacronym{aid}{AID}{Average Inference Duration}
\newacronym{iot}{IoT}{Internet of Things}
\newacronym{lm}{LM}{language model}
\newacronym{ml}{ML}{Machine Learning}
\newacronym{dl}{DL}{Deep Learning}
\newacronym{wot}{WoT}{Web of Things}
\newacronym{saref}{SAREF}{Smart Applications REFerence Ontology}
\newacronym{ogc}{OGC}{Open Geospatial Consortium}
\newacronym{fhir}{FHIR}{Fast Healthcare Interoperability Resources}
\newacronym{mim}{MIM}{Minimum Interoperability Mechanisms}
\newacronym{oasc}{OASC}{Open Agile Smart City}
\newacronym{opcua}{OPC UA}{Open Platform Communications Unified Architecture}
\newacronym{shacl}{SHACL}{Shapes Constraint Language}
\newacronym{diim}{DIIM}{Data and Information Interoperability Model}
\newacronym{mda}{MDA}{Model-Driven Architecture}
\newacronym{swn}{SWN}{Smart Water Networks}
\newacronym{bimiotdi}{BIM-IoTDI}{BIM-IoT Data Integration}
\newacronym{bim}{BIM}{Building Information Modeling}
\newacronym{tfidf}{TF-IDF}{Term Frequency–Inverse Document Frequency}

\definecolor{light-green}{HTML}{7cf280}
\definecolor{light-yellow}{HTML}{ffe987}
\definecolor{light-red}{HTML}{fc9595}

\title{Small Language Models for Smart Data Model Classification at the Edge: A Cost-Aware Hybrid Approach}

\author[1]{Cristian Martella}
\author[1]{Angelo Martella\corref{cor1}}
\author[1]{Antonella Longo}
\author[1]{Motaz Saad}
\cortext[cor1]{Corresponding author}
\affiliation[1]{organization={Dept. of Engineering for Innovation, University of Salento},
            addressline={Via Per Monteroni},
            city={Lecce},
            postcode={73100},
            country={Italy}}

\date{}

\usepackage{caption}
\usepackage{newfloat}
\usepackage{tcolorbox}
\usepackage[breaklinks=true]{hyperref}
\DeclareFloatingEnvironment[
  fileext=lob,
  listname={List of Boxes},
  name=Listing,
  placement=htp,
]{BOX}

\begin{document}
\begin{sloppypar}
\small
\begin{abstract}
The rapid proliferation of heterogeneous data sources within the \gls{iot} across domains such as smart cities, energy management, and environmental monitoring necessitates efficient and scalable data standardization methods. Effective classification of \glspl{sdm} is essential for facilitating interoperability. However, existing approaches are often limited by high resource consumption and lack applicability in edge environments with constrained computational capabilities. Aiming to bridge this gap, the proposed study evaluates the performance of lightweight open-source \glspl{lm} to resolve an input data entity against its corresponding best fitting \gls{sdm} representation under resource-constrained conditions. It systematically benchmarks a diverse array of models, including \gls{gp}, \gls{rs}, and \gls{cs} architectures, across multiple domain-specific datasets. Addressing the current omission of lightweight, resource-efficient solutions in the literature, the investigation provides significant and valuable insights into model selection, task formulation, and deployment strategies that optimize accuracy and efficiency. A complementary experiment also compares the surveyed \glspl{llm} against two near-zero-cost similarity baselines (\gls{tfidf} and a lightweight sentence encoder) on the same task, providing a strong reference point for interpreting the practical value of \gls{llm}-based classification on edge platforms.
\end{abstract}


\maketitle

\vspace*{-1\baselineskip}
\section{Introduction}
\label{s:introduction}
In the rapidly evolving landscape of the \gls{iot}, the proliferation of heterogeneous data sources across domains such as smart cities, energy management, and environmental monitoring necessitates effective and scalable data integration solutions. \Glspl{sdm} play a critical role in the data standardization landscape, but their effective applicability is hindered by traditional approaches to entity classification and schema matching, which pose significant challenges related to semantic heterogeneity and resource constraints, especially in edge-fog environments. This paper aims to systematically evaluate the feasibility of lightweight open-source \glspl{lm} for structured data classification tasks within resource-constrained \gls{iot} ecosystems. To achieve this, the study explores a range of \gls{gp}, \gls{rs}, and \gls{cs} models, benchmarking their performance on diverse \Glspl{sdm} drawn from multiple domains.

\begin{figure}
\centering
\resizebox{\linewidth}{!}{%
\begin{tikzpicture}[
    device/.style={rectangle, rounded corners, draw=#1!70, fill=#1!10,
                   thick, minimum height=0.8cm, minimum width=3cm,
                   drop shadow, font=\footnotesize},
    repository/.style={cylinder, cylinder uses custom fill,
                      cylinder body fill=purple!15, cylinder end fill=purple!25,
                      draw=purple!70, very thick, minimum height=3cm,
                      minimum width=3.5cm, drop shadow, font=\normalsize\bfseries,
                      shape aspect=0.25, align=center},
    arrow/.style={-{Stealth[length=2.5mm]}, thick, draw=gray!60},
    databox/.style={font=\ttfamily\tiny, align=left}
]
\node[device=blue] (d1) at (0, 3.5)  {Vendor A -- Sensor 1};
\node[databox, below=0.05cm of d1, text=blue!70!black]   {temp: 22.5\\client: "John"};
\node[device=blue] (d2) at (0, 1.8)  {Vendor A -- Sensor 2};
\node[databox, below=0.05cm of d2, text=blue!70!black]   {moisture: 65\\user: "Alice"};
\node[device=green!80!black] (d3) at (8, 3.5) {Vendor B -- Device 1};
\node[databox, below=0.05cm of d3, text=green!50!black]  {temperature: 21.8\\customer: "John"};
\node[device=green!80!black] (d4) at (8, 1.8) {Vendor B -- Device 2};
\node[databox, below=0.05cm of d4, text=green!50!black]  {humidity: 68\\clientName: "Bob"};
\node[device=orange] (d5) at (0, 0)   {Vendor C -- Monitor 1};
\node[databox, below=0.05cm of d5, text=orange!60!black] {ambient\_temp: 23\\cust: "Mary"};
\node[device=orange] (d6) at (0, -1.7){Vendor C -- Monitor 2};
\node[databox, below=0.05cm of d6, text=orange!60!black] {relHumidity: 70\\buyer: "Jane"};
\node[device=red] (d7) at (8, 0)   {Vendor D -- Station 1};
\node[databox, below=0.05cm of d7, text=red!70!black]    {tmp: 22.1\\person: "Tom"};
\node[device=red] (d8) at (8, -1.7){Vendor D -- Station 2};
\node[databox, below=0.05cm of d8, text=red!70!black]    {hum: 72\\contact: "Eve"};
\node[repository] (repo) at (4, 0.9) {Unified\\Repository};
\foreach \src in {d1,d2,d3,d4,d5,d6,d7,d8} {
    \draw[arrow] (\src) -- (repo);
}
\node[rectangle, draw=red!70, fill=yellow!10, thick, rounded corners,
      minimum width=9cm, minimum height=1cm, align=center, font=\small]
      at (4, -3.2) {\textbf{Interoperability Challenge:}\\
                    Same semantic concepts with different syntactic representations};
\end{tikzpicture}%
}
\caption{Heterogeneous data integration challenge in \gls{iot} ecosystems with multiple vendors}
\label{fig:iot-heterogeneity}
\end{figure}
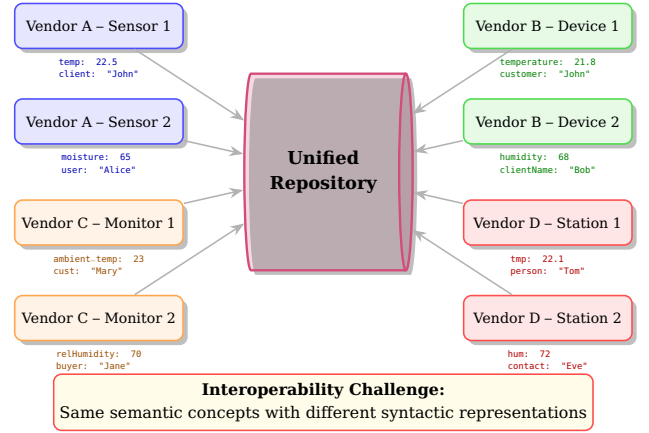

Figure~\ref{fig:iot-heterogeneity} illustrates a representative scenario where environmental sensors and monitoring stations from four different vendors transmit data to a centralized repository. Despite measuring identical physical phenomena (such as temperature and humidity) and tracking the same types of entities (\textit{customers} or \textit{users}), each vendor employs distinct attribute naming conventions and data schemas. For instance, temperature readings are variously encoded as \textit{temperature}, \textit{temp}, \textit{ambientTemp}, and \textit{tmp}; while customer identifiers appear as \textit{client}, \textit{customer}, \textit{cust}, and \textit{person}. This syntactic heterogeneity, combined with the absence of standardized metadata vocabularies, creates substantial barriers to semantic interoperability and automated data fusion, underscoring the need for sophisticated ontology alignment and schema matching techniques in modern \gls{iot} platforms.

In this context, \gls{er} is the process of identifying and matching records that refer to the same real-world entity across different datasets or data streams. The role of \Gls{er} in data model standardization is to identify and link semantically equivalent attributes and records across heterogeneous data models, thereby enabling interoperability and consistent representation of entities across diverse systems and domains.

In the literature, various approaches exist to address \Gls{er} that can be organized into 4 main research lines based on rules, ontologies, \gls{ml} and \gls{dl} techniques, and \gls{llm} technology, respectively. The most recent studies on \glspl{llm} appear to be promising, although they are often resource- and cost intensive.
Furthermore, \Gls{er} and data model standardization primarily focus on large-scale cloud-based models or traditional heuristic approaches, leaving a significant gap in the investigation of lightweight models suitable for edge-fog deployments. As a consequence, there is limited understanding of how different model architectures, parameter sizes, and task formulations influence classification accuracy in resource constrained scenarios. This work seeks to address these gaps by conducting a comprehensive benchmarking of 26 lightweight models under real-world constraints, providing practical insights and actionable recommendations for deploying automated \Gls{sdm} classification systems in heterogeneous \gls{iot} ecosystems. In doing so, it aims to advance the state of knowledge toward scalable, efficient, and domain-robust data integration methodologies.

It is important to position this work relative to established deterministic approaches to \gls{iot} sensor onboarding. Standards such as W3C \gls{wot} Thing Descriptions\footnote{\scriptsize{\url{https://www.w3.org/TR/wot-thing-description/}}}, NGSI-LD explicit context definitions, and declarative transformation languages such as JSONata\footnote{\scriptsize{\url{https://jsonata.org/}}} provide guaranteed precision and sub-millisecond latency when device schemas are pre-registered or hand-authored mapping rules exist. For known and well-documented device types, these approaches are the natural and preferred choice. However, they fundamentally require upfront manual engineering: a domain expert must author and maintain a mapping rule or Thing Description for each device type. The multi-vendor heterogeneity illustrated in Figure~\ref{fig:iot-heterogeneity}, where four vendors independently encode the same physical phenomenon using incompatible naming conventions, represents the pervasive cold-start condition encountered when integrating third-party, legacy, or previously unseen devices for which no pre-existing mapping exists. 

In this context, \gls{llm}-based \gls{er} does not compete with deterministic methods; rather, it addresses the prior step of automated entity discovery, classifying incoming data entities against a standardized \gls{sdm} taxonomy to bootstrap the onboarding process and reduce the manual authoring effort required before deterministic integration can be established.

A common approach to comparing pieces of text, such as an incoming data entity's JSON keys and candidate taxonomy entries, is to first convert each text into a numerical vector (a process known as text vectorization) and then measure the cosine of the angle between the resulting vectors. Two widely used vectorization schemes are compared in this work as baselines. The first, \gls{tfidf}, represents each text by the (weighted) frequency of its constituent tokens; it captures lexical overlap and requires no training. The second vectorization
schema embeds the same text with a lightweight sentence encoder (\texttt{all-MiniLM-L6-v2}), producing dense, semantically-grounded vectors that can generalise beyond exact token matches. By benchmarking the surveyed \glspl{llm} against both a lexical and a semantic baseline on the same classification task, the study provides a concrete reference point for assessing when the additional cost of an \gls{llm} is justified.

This research illustrates the trade-offs between model complexity, accuracy, and inference efficiency, by systematically analyzing the performance of various lightweight models across multiple classification strategies and domain-specific datasets. Critical factors such as task formulation, model specialization, and parameter scaling are investigated, challenging conventional assumptions about the supremacy of certain architectures for structured data classification. Furthermore, this work highlights significant research gaps, including the limited focus on resource-efficient models for dynamic, large-scale data streams. Addressing this issue, the study provides foundational insights that can guide practitioners and researchers in designing scalable, accurate, and computationally feasible models for semantic data integration in \gls{iot} environments, ultimately contributing to more interoperable and intelligent urban, energy, and environmental ecosystems.

Following this introduction, Section~\ref{s:bg_related_work} illustrates the landscape of recent related works, Section~\ref{s:methodology} describes the methodology adopted for the purposes of this research, Section~\ref{s:results_discussion} analyzes the results of the experiments, Section~\ref{s:final_remarks} provides some final insights and considerations for designers and practitioners, and Section~\ref{s:conclusion} concludes the article.

\vspace*{-1\baselineskip}
\section{Background and Related Work}
\label{s:bg_related_work}

This section reviews the state of the art in \Gls{dm} standardization, classification methodologies, and \gls{lm} applications in structured data tasks. Schema and attribute name resolution is a fundamental task in data integration, enabling systems to identify semantically equivalent attributes across heterogeneous \Glspl{dm} (e.g., mapping "cust\_id" to "customerID"). Traditional approaches rely on lexical similarity, rule-based heuristics, or supervised \gls{ml}, but these methods often struggle with semantic heterogeneity and lack scalability.

Recent research explores the use of \Glspl{llm} for schema and attribute-level matching, leveraging their contextual and semantic understanding.

\vspace*{-.75\baselineskip}
\subsection{Data Model Standardization}
\label{ss:data_model_standardization}
Entity \gls{dm} standardization involves creating consistent, interoperable frameworks for representing and exchanging data about entities across different systems and domains. Such a standardization is achieved by defining and maintaining the interoperable data entity representation frameworks and by designing and implementing \Gls{er} flows.
In this sense, \Gls{er} is the process of identifying and linking records that refer to the same real-world entity across different datasets.

Entity \gls{dm} standardization has evolved along various distinct methodological trajectories optimized for different application contexts. Ontology-based semantic approaches using \gls{saref}~\cite{Bjrnskov2023}, Brick~\cite{Pereira2025}, and \gls{ogc} standards~\cite{Huang2022} dominate in \gls{iot}, building, and smart city applications where semantic richness and flexible integration of heterogeneous data sources are paramount, achieving demonstrated consistent efficiency gains in building control deployment but facing challenges with traditional modeling tool compatibility~\cite{Bjrnskov2023}. Schema-based standards like \gls{fhir}~\cite{Tabari2024} and FIWARE’s NGSI-LD~\cite{Bauer2022} prevail in healthcare and cross-domain platforms where formal validation and broad adoption are critical. In particular, NGSI-LD is adopted as a \gls{mim} by the \gls{oasc} network and \gls{fhir} being recognized as a highly promising healthcare interoperability standard, though both NGSI-LD and \gls{fhir} face ongoing challenges in data integration, performance, and scalability. Industrial automation employs specialized standards including \gls{opcua} and Asset Administration Shell~\cite{Schlemitz2024}, where no single standard covers all use cases~\cite{Miny2023}, necessitating integration strategies that combine multiple standards.

The effectiveness of each approach depends fundamentally on domain requirements rather than inherent technical superiority. Federated and modular architectures achieve scalability through loose coupling and reusability~\cite{Ahle2022}, while standardized APIs and semantic alignment enable interoperability improvements. Anyway, universal agreement on standards remains challenging and interoperability gaps persist at domain integration points~\cite{Bauer2022}. Healthcare applications face critical needs for \gls{dm} convergence across diverse data modalities~\cite{Cremonesi2023}, smart cities struggle with proprietary technology integration~\cite{Bauer2022}, and construction faces challenges from the proliferation of different approaches~\cite{Nuyts2024}, with \gls{shacl} emerging as the most standardized Linked Data validation method.
Currently, many initiatives to \gls{dm} standardization are in going and among them the most significant for this study is \Glspl{sdm}.

The \Glspl{sdm} initiative is a collaborative program that provides harmonised representation formats and semantics for applications to consume and publish data across multiple industrial domains. The initiative is jointly led by FIWARE Foundation, IUDX (India Urban Data Exchange), TM Forum, and \gls{oasc}, with the goal of creating interoperable and replicable smart solutions. In particular, the \Glspl{sdm} initiative provides compatible, domain-specific \glspl{dm} to facilitate data sharing and understanding across different applications \cite{bauer2022fiware}, by promoting NGSI-LD standard format. In effect, NGSI-LD has been standardized by the ETSI Industry Specification Group on Context Information Management.

FIWARE models are designed to be generic and reusable across different domains, while ontologies are often tailored to specific applications or industries. This means that FIWARE models may not capture all the nuances and complexities of a particular domain, but they provide a common foundation for data integration and exchange.

The \Gls{dm} datasets are hosted on GitHub\footnote{\scriptsize{\url{https://github.com/smart-data-models}}} into 101 repositories organized hierarchically. The structure comprises an umbrella repository that contains all \Glspl{dm} across domain collections for each industry sector with submodules linking to subjects, and subject-specific vertical sub-repositories that contain the corresponding \Glspl{dm}.


The \Glspl{sdm} are structured as JSON-LD (JSON for Linking Data) documents following JSON Schema specifications with accompanying documentation and context specification. Each model is well documented by providing: (a) A JSON-LD example demonstrating the complete structure and proper usage; (b) Semantic annotations using standard vocabularies (schema.org, FIWARE, etc.); (c) Property definitions with data types, constraints, and relationships; (d) Context information linking to external vocabularies.

The models follow a driven-by-implementation approach where specifications are considered stable once enough end-user organizations have validated them in practice. Each model is versioned annually with the \textit{schemaVersion} parameter denoting the version.

FIWARE models, alongside NGSI-LD, have been adopted by the \gls{oasc} network as \gls{mim}.

Several other initiatives aim to address interoperability challenges using ontologies and semantic web technologies. Among the domain-specific ontologies, it is worth mentioning the proposal of a \gls{diim} that combines semantic web technologies and \gls{mda} to facilitate syntactic and semantic interoperability in \gls{iot}-enabled \gls{swn}~\cite{Singh2022}. In smart buildings domain, \gls{bimiotdi}~\cite{Eneyew2022} enables semantic interoperability between \gls{bim} and \gls{iot} data sources.

Other relevant initiatives include the following: IDSA Information Model\footnote{\scriptsize{\url{https://github.com/International-Data-SpacesAssociation/InformationModel}}}, \gls{saref} ontology\footnote{\scriptsize{\url{https://saref.etsi.org/}}}, Blockchain in Transport Alliance (BiTAS)\footnote{\scriptsize{\url{https://www.bita.studio/}}}, DCSA Interface for Track \& Trace\footnote{\scriptsize{\url{https://dcsa.org/standards/track-trace/}}}, United Nations Centre for Trade Facilitation and Electronic Business (UN/CEFACT) model, IPSO Smart Objects (OMA SpecWorks)\footnote{\scriptsize{\url{https://omaspecworks.org/develop-with-oma-specworks/ipso-smart-objects/}}}.

Recent studies focus on the integration of \Glspl{llm} with ontologies, showcasing new opportunities for enhancing ontology learning and knowledge extraction \cite{Peng2024,Bendiken2024}. \glspl{llm} can be used to automate and improve the ontology and information model learning process, offering information on effective methodologies for future developments in this field.

\vspace*{-.75\baselineskip}
\subsection{Entity Resolution}
\label{ss:entity_resolution}
The \gls{er} process is a well-known method in the literature that aims to enhance data quality by uniformly representing entities in datasets or streams~\cite{Martella2024,Christophides2020}. It involves a series of steps (primarily blocking, matching, and clustering) to identify matching pairs of entity descriptions corresponding to real-world objects, also called \textit{duplicates}. The steps in a typical \gls{er} process are described in the following, along with significant state-of-the-art works.

The \textit{Blocking} and \textit{Block Processing} stage helps handle large volumes of data by separating input into blocks based on similarity criteria. Dynamic indexing techniques, such as Similarity-aware Index~\cite{Christen2009} and DySimII~\cite{Ramadan2013}, update these blocks based on entity descriptions obtained through queries or streams of entities, which can be interpreted as sequences of queries.

\textit{Matching} involves forwarding entity descriptions to a matching function based on binary similarity. Dynamic Matching approaches like QuERy~\cite{Altwaijry2015} and QDA~\cite{Altwaijry2017} are not designed for incremental execution and do not support functions considering more than two values. In other words, such approaches are not designed to deal with data streamed dynamically by diverse data sources, but rather perform best with static datasets. BrewER~\cite{Simonini2022,Zecchini2022} can be employed within each block to reduce similarity comparisons and feed the query tree with resolved entities, dynamically and on demand. This approach is preferred due to its reliability, effectiveness, and efficiency.

In the \textit{Clustering} stage, duplicates are inferred by grouping them within clusters, improving the similarity function used in the Matching stage. This results in a set of entity clusters, each described by features of a real world object. Incremental Correlation Clustering is the leading approach to Dynamic Clustering~\cite{Gruenheid2014}, capable of performing updates on clusters without prior knowledge of the number of clusters involved.

Given its complexity, traditionally \gls{er} is a resource intensive task that requires significant computing power to achieve valid performance~\cite{Christophides2020}. Moreover, parallelization strategies for \gls{er} tasks are adopted to scale to large volumes of data. Yet, such strategies are not feasible to support incremental or streaming \gls{er}~\cite{Gazzarri2021}, i.e. \gls{er} performed on live data streamed by dynamic data sources such as \gls{iot} devices.
Constraints related to computing resources or time have been overlooked by research but are becoming increasingly influent as edge-cloud continuum paradigm takes the stage. The main goal of such architectures is to shift high-performance computing tasks towards the edge of the Internet, closer to where data is produced or consumed~\cite{Martella2024}. In this way, network traffic and computing load balance are optimized, while offering opportunities in terms of data sovereignty constraints and privacy preservation.
Budget-aware \gls{er} are intended to provide the best possible partial solution in resource constrained environments~\cite{Christophides2020}.

Considering the recent outbreak of AI applications, current approaches to \gls{er} can be organized into four main categories, discussed in the following. Specifically, \gls{er} techniques can be based on rules, ontologies, \gls{ml} and \gls{dl}, and \Gls{llm} approaches.

\subsubsection{Rule-based Approaches}
\label{sss:rule_based_approaches}
Beyond schema-based standards, deterministic sensor onboarding pipelines represent the established baseline for \gls{iot} data integration in production deployments. The W3C \gls{wot} architecture defines a Thing Description (TD) format that allows devices to self-describe their capabilities and semantic properties, eliminating runtime schema inference when TDs are available. NGSI-LD context definitions allow platform operators to pre-register entity types and attribute mappings, enabling 100\% precision classification at millisecond latency. Declarative transformation languages such as JSONata provide rule-based field remapping that is both auditable and entirely deterministic. These approaches excel in controlled environments where device schemas are stable and known in advance. Their principal limitation is that they require prior investment in manual rule authoring, which becomes impractical at scale when integrating diverse third-party or legacy devices with heterogeneous and undocumented attribute naming conventions. 

\gls{llm}-based \gls{er} is therefore best understood as complementary to these deterministic pipelines: it targets the cold-start phase of device onboarding where no pre-existing mapping exists, automating initial entity classification to reduce the manual effort required before deterministic rules can be established.

\subsubsection{Ontology-based Approaches and Semantic Layer Integration}
\label{sss:ontology_based_approaches}
The field of data \Gls{er} has made impressive strides thanks to the use of ontology-based approaches. These methods leverage ontologies to address \Gls{er} to improve accuracy, clarity, and depth of understanding in matching entities. Recent research has shown how ontologies and related semantic frameworks can tackle the challenges that come with \Gls{er} tasks.
Ontology-based approaches to data \Gls{er} in recent years can be clustered into three distinct methodological paradigms, each optimized for different operational contexts.

Embedding-based approaches employ neural networks and statistical learning to automatically align entities across knowledge graphs, with methods like BERT\_INT achieving dominance in datasets with high factual similarity \cite{Fanourakis2023,Sanjak2023}. These approaches offer strong automation benefits through \gls{ml} but face limitations in computational complexity requiring negative sampling for dense datasets, strict assumptions limiting domain transferability, and limited interpretability requiring additional manual effort for translatable results \cite{Fanourakis2023}.

Knowledge-enhanced approaches incorporate external background knowledge via factual queries, structure-based methods, statistical/neural techniques, and logic-based reasoning. They yield higher recall and F-measure when domain-specific resources are available, but scalability is limited by resource size and linking requires manual effort \cite{Portisch2022}.

Domain~-specific semantic integration approaches prioritize interoperability through standardized frameworks like RDF/OWL, enabling SPARQL queries and cross-system integration while facilitating semantic reasoning and structured data interpretation \cite{DAuria2023}. At the same time, they face challenges in scalability related to developing fully interoperable semantic models, reliance on commercial software, and requirements for specialized expertise across multiple fields \cite{Celeste2022,Belani2025}.

Essentially, ontology-based and semantic-layer approaches provide additional robustness by mapping schema elements to knowledge graphs or controlled vocabularies \cite{molnar2025kgraphrag}.

\subsubsection{Machine Learning- and Deep Learning-based Approaches}
\label{sss:machine_deep_learning_based_approaches}
\gls{ml} \gls{er} has evolved to address complex data integration challenges, with recent approaches focusing on automation, knowledge integration, and reduced human intervention. 
\gls{ml}-based approaches to \gls{er} typically employ supervised or unsupervised algorithms to identify and match records that refer to the same real-world entity across different datasets. These methods often use feature engineering to extract similarities from entity attributes before applying classification or clustering algorithms to determine matches. 

On the contrary, ontology-based approaches leverage structured knowledge representations and semantic relationships to resolve entities by mapping them to standardized concepts in domain-specific ontologies, using logical reasoning and semantic similarity measures.

Studies in literature span multiple \gls{er} subtasks related to matching \cite{Leone2022}, blocking (which reduces computational complexity by identifying candidate pairs before matching) \cite{Thirumuruganathan2021,Javdani2019,Kasai2019}, and the broader entity alignment problem \cite{DiCicco2019,Dou2024}.

Recent research reveals three primary hybridization strategies for ML and ontologies: learning enhanced ontologies, semantic data mining, and integrated learning reasoning systems \cite{Ghidalia2024,Papadakis2021}. 


\gls{dl} consistently outperforms traditional methods on textual and dirty data but shows limited advantages on clean, structured data \cite{Mudgal2018}.

\subsubsection{LLM-Based Approaches}
\label{sss:llm_based_approaches}
Approaches to \gls{er} based on \gls{llm} represent a paradigm shift from traditional methods by leveraging the advanced natural language understanding and reasoning capabilities of \glspl{llm} to identify and match entities across datasets \cite{Huang2026}. Unlike traditional \gls{ml}-based approaches, LLM-based methods can perform ER through natural language instructions, zero-shot or few-shot prompting, and contextual reasoning without extensive domain-specific training. They also differ from ontology-based approaches by utilizing the implicit world knowledge embedded in pretrained \glspl{lm} to understand semantic relationships and handle ambiguous or incomplete data more flexibly.

Recent LLM-based \gls{er} methodologies demonstrate substantial performance gains with significant accuracy improvements~\cite{Fu2025,Low2024,Ding2024} across diverse tasks including entity matching~\cite{Wang2024,Dou2026}, alignment~\cite{Chen2024}, linking~\cite{Xin2025}, and deduplication~\cite{Li2024}. Key innovations explore varying-size \glspl{llm} for initial filtering and final decision-making of data entities ~\cite{Fu2025}, hybrid methods that integrate knowledge graphs with \glspl{llm}, and uncertainty-reduction frameworks that optimize question selection ~\cite{Li2024}.

While recent LLM-based \gls{er} approaches demonstrate substantial performance improvements across matching, alignment, and linking tasks~\cite{Fu2025,Wang2024,Chen2024,Xin2025}, they predominantly focus on large-scale models (30B+ parameters) deployed via cloud APIs, with emphasis on accuracy gains and API cost optimization through selective deployment strategies~\cite{Fan2024,Li2024}. This literature exhibits three critical gaps for \gls{iot} scenarios. First, \textit{resource constraints}: existing work overlooks deployment feasibility in edge-fog environments where computational resources are severely limited, cloud connectivity unreliable, and energy efficiency critical~\cite{Martella2024}. Second, \textit{domain specificity}: evaluations focus on general entity matching rather than hierarchical, domain-specific taxonomies like \Glspl{sdm} requiring multi-level classification across standardized \gls{iot} verticals. Third, \textit{model specialization}: systematic comparisons of \Gls{gp}, \Gls{rs}, and \Gls{cs} models within the lightweight regime ($\leq$4B parameters) are absent, leaving practitioners without guidance on which architectural approaches best balance performance and efficiency for structured classification tasks. Furthermore, the assumption that providing taxonomies uniformly improves classification remains untested in the context of \gls{er}, despite potential cognitive overhead from large option sets.

\vspace*{-1\baselineskip}
\section{Methodology}
\label{s:methodology}

This section describes our experimental design, including the dataset composition, model selection rationale, prompt engineering approach, and evaluation framework.
Before proceeding with the discussion, it is fundamental to identify gaps in the literature regarding \Gls{er}.

Following the review proposed in section \ref{ss:entity_resolution}, some gaps in the literature become evident in terms of the ability of current \Gls{er} approaches to handle large-scale and dynamic data streams in resource constrained environments. In this paper, the goal is to benchmark state-of-the-art lightweight \Glspl{llm} to evaluate their feasibility as core matching components of innovative \Gls{er} pipelines. The study focuses specifically on lightweight \Glspl{llm} to address the requirements posed by resource constrained environments, specifically in terms of computing power and inference time. Performing a reliable \Gls{er} pipeline in low-power computing infrastructure (typically on devices in the edge of the Internet) fosters data sovereignty and helps responding to privacy-preservation requirements. Nevertheless, this also poses challenges in terms of balance between accuracy of the model vs. the available budget in terms of resources.

\vspace*{-.75\baselineskip}
\subsection{System Architecture and Classification Pipeline}
\label{ss:system_architecture_methodology}

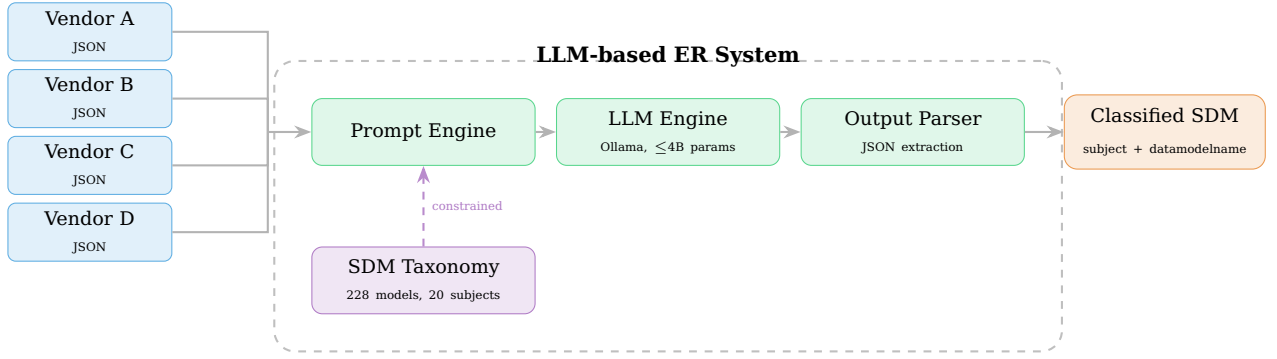
\begin{figure*}[!ht]
\centering
\resizebox{\linewidth}{!}{%
\begin{tikzpicture}[
    source/.style={rectangle, rounded corners=3pt, draw=iotblue!80, fill=iotblue!15,
                   text width=2cm, align=center, minimum height=0.7cm, font=\footnotesize},
    comp/.style={rectangle, rounded corners=4pt, draw=llmgreen!80, fill=llmgreen!15,
                 text width=2.8cm, align=center, minimum height=0.9cm, font=\footnotesize},
    tax/.style={rectangle, rounded corners=4pt, draw=ragpurple!80, fill=ragpurple!15,
                text width=2.8cm, align=center, minimum height=0.9cm, font=\footnotesize},
    outbox/.style={rectangle, rounded corners=5pt, draw=standardorange!80, fill=standardorange!15,
                text width=2.5cm, align=center, minimum height=1cm, font=\footnotesize},
    sysbox/.style={rectangle, rounded corners=8pt, draw=gray!50, 
    dashed, thick},
    arr/.style={->, >=Stealth, thick, draw=gray!60},
    darr/.style={->, >=Stealth, thick, draw=ragpurple!70, dashed},
    lbl/.style={font=\small\bfseries, align=center}
]

\node[source] (va) at (0,  1.35) {Vendor A\\{\tiny JSON}};
\node[source] (vb) at (0,  0.45) {Vendor B\\{\tiny JSON}};
\node[source] (vc) at (0, -0.45) {Vendor C\\{\tiny JSON}};
\node[source] (vd) at (0, -1.35) {Vendor D\\{\tiny JSON}};
\node[lbl] at (0, 2.1) {IoT Data Sources};

\node[comp] (prompt) at (4.5,  0.0) {Prompt Engine};
\node[tax]  (tax)    at (4.5, -2.0) {SDM Taxonomy\\{\tiny 228 models, 20 subjects}};
\node[comp] (llm)    at (7.8,  0.0) {LLM Engine\\{\tiny Ollama, $\leq$4B params}};
\node[comp] (parser) at (11.1, 0.0) {Output Parser\\{\tiny JSON extraction}};

\node[sysbox, fit=(prompt)(tax)(llm)(parser), inner sep=0.5cm] (sys) {};
\node[lbl] at (sys.north) [above=-0.2cm] {LLM-based ER System};

\node[outbox] (output) at (14.5, 0.0) {Classified SDM\\[2pt]{\tiny subject + datamodelname}};

\coordinate (collect) at (2.4, 0.0);
\draw[thick, gray!60] (va.east) -| (collect);
\draw[thick, gray!60] (vb.east) -| (collect);
\draw[thick, gray!60] (vc.east) -| (collect);
\draw[thick, gray!60] (vd.east) -| (collect);
\draw[arr] (collect) -- (prompt.west);

\draw[darr] (tax.north) -- (prompt.south)
    node[midway, right, font=\tiny, text=ragpurple!80] {constrained};

\draw[arr] (prompt) -- (llm);
\draw[arr] (llm)    -- (parser);

\draw[arr] (parser.east) -- (output.west);

\end{tikzpicture}%
}
\caption{System architecture of the LLM-based \gls{er} system for \gls{sdm} classification in resource-constrained \gls{iot} environments. The \gls{sdm} taxonomy is used only in the constrained classification path (dashed arrow).}
\label{fig:architecture}
\end{figure*}

\begin{figure}[!ht]
\centering
\resizebox{\linewidth}{!}{%
\begin{tikzpicture}[
    io/.style={rectangle, rounded corners=5pt, draw=iotblue!80, fill=iotblue!15,
               text width=4cm, align=center, minimum height=1cm, font=\small\bfseries},
    prompt/.style={rectangle, rounded corners=4pt, draw=ragpurple!80, fill=ragpurple!15,
                   text width=4cm, align=center, minimum height=1cm, font=\small},
    proc/.style={rectangle, rounded corners=4pt, draw=llmgreen!80, fill=llmgreen!15,
                 text width=4cm, align=center, minimum height=1cm, font=\small},
    outbox/.style={rectangle, rounded corners=4pt, draw=standardorange!80, fill=standardorange!15,
                text width=4cm, align=center, minimum height=1cm, font=\small},
    arr/.style={->, >=Stealth, thick, draw=gray!60},
    lbl/.style={font=\small\bfseries}
]

\node[io] (input) at (0,0) {IoT Data Entity\\(JSON)};

\node[lbl] at (-3.5,-1.1) {Constrained Path};
\node[lbl] at ( 3.5,-1.1) {Unconstrained Path};

\node[prompt] (cprompt) at (-3.5,-2.4) {Prompt Construction\\{\footnotesize entity + SDM taxonomy\\(228 models, 20 subjects)}};
\node[proc]   (cllm)    at (-3.5,-4.2) {LLM Inference\\{\footnotesize via Ollama (temp=0.4)}};
\node[proc]   (cparse)  at (-3.5,-5.8) {Output Parsing\\{\footnotesize JSON extraction}};
\node[outbox] (cout)    at (-3.5,-7.4) {Classified SDM\\{\footnotesize \texttt{\{subject, datamodelname\}}}};

\node[prompt] (uprompt) at (3.5,-2.4) {Prompt Construction\\{\footnotesize entity only}};
\node[proc]   (ullm)    at (3.5,-4.2) {LLM Inference\\{\footnotesize via Ollama (temp=0.4)}};
\node[proc]   (uparse)  at (3.5,-5.8) {Output Parsing\\{\footnotesize JSON extraction}};
\node[outbox] (uout)    at (3.5,-7.4) {Classified SDM\\{\footnotesize \texttt{\{datamodelname,}}\\\footnotesize\texttt{synonym\_1/2/3\}}};

\draw[thick, gray!60] (input.south) -- ++(0,-0.5) coordinate (fork);
\draw[arr] (fork) -| (cprompt.north);
\draw[arr] (fork) -| (uprompt.north);

\draw[arr] (cprompt) -- (cllm);
\draw[arr] (cllm)    -- (cparse);
\draw[arr] (cparse)  -- (cout);

\draw[arr] (uprompt) -- (ullm);
\draw[arr] (ullm)    -- (uparse);
\draw[arr] (uparse)  -- (uout);

\end{tikzpicture}%
}
\caption{Overview of the LLM-based \gls{sdm} classification pipeline with constrained and unconstrained paths.}
\label{fig:pipeline}
\end{figure}
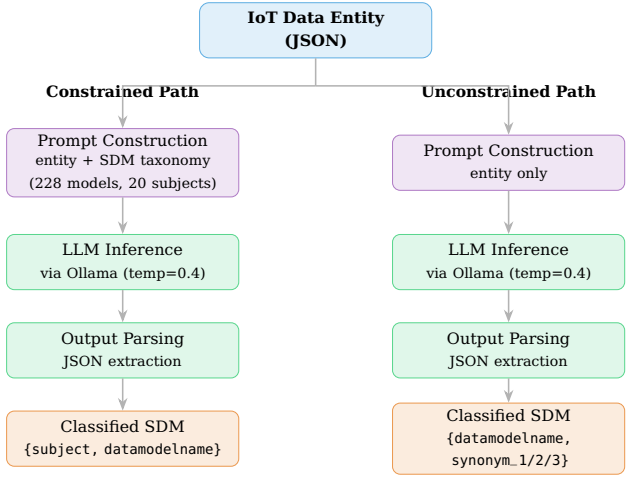

To clarify how the benchmarked \glspl{llm} are embedded in an end-to-end \gls{er} workflow, this subsection describes the proposed system architecture and the classification pipeline adopted in the experimental study. Figure~\ref{fig:architecture} reports the high-level architecture, while Figure~\ref{fig:pipeline} details the per-entity processing pipeline along the two classification paths investigated in this work (\textit{constrained} and \textit{unconstrained}, see Section~\ref{ss:classification-framework}).

\paragraph{System Architecture} Heterogeneous \gls{iot} data sources (e.g., JSON payloads emitted by different vendors) feed the LLM-based \gls{er} system. Inside the system (see Figure~\ref{fig:architecture}), four logical components cooperate: (i) the \textit{Prompt Engine}, which assembles the input entity into a task-specific prompt and, in the constrained path only, injects the \gls{sdm} taxonomy (228 data models grouped under 20 subjects) as auxiliary context; (ii) the \textit{LLM Engine}, which executes inference through the Ollama runtime using lightweight open-source models with at most $4$B parameters; (iii) the \textit{Output Parser}, which extracts structured fields from the raw model response; and (iv) the resulting \textit{Classified \gls{sdm}}, encoded as a $\langle$\textit{subject}, \textit{datamodelname}$\rangle$ pair. The dashed arrow from the \gls{sdm} taxonomy to the Prompt Engine highlights that the taxonomy is consumed exclusively in the constrained classification path.

\paragraph{Classification Pipeline} For each input \gls{iot} data entity, the pipeline (see Figure~\ref{fig:pipeline}) branches into two parallel paths to enable a fair comparison of taxonomy-aware and taxonomy-free classification. Both paths share the same four stages: \textit{Prompt Construction}, \textit{LLM Inference}, \textit{Output Parsing}, and \textit{Classified \gls{sdm}}. In the \textbf{constrained path}, prompt construction concatenates the input entity with the full \gls{sdm} taxonomy, asking the model to select the best matching $\langle$\textit{subject}, \textit{datamodelname}$\rangle$ pair from the provided options. In the \textbf{unconstrained path}, prompt construction includes only the input entity, asking the model to produce a \textit{datamodelname} together with up to three synonyms, with no taxonomy reference. LLM inference is executed in both cases through the Ollama API with temperature set to $0.4$. Output parsing applies JSON-mode decoding followed by Python-level key extraction, as detailed in Section~\ref{ss:parsing-pipeline}. The two paths terminate in structurally different outputs: the constrained path returns a taxonomy-validated $\langle$\textit{subject}, \textit{datamodelname}$\rangle$ pair, whereas the unconstrained path returns a free-form \textit{datamodelname} accompanied by candidate synonyms suitable for downstream lexical or semantic matching.

\vspace*{-.75\baselineskip}
\subsection{Target Smart Data Models}
\label{ss:target-smart-data-models}

FIWARE's \Glspl{sdm} aim to establish a common foundation for data integration and exchange. To this end, they are designed to be generic and cross-domain.

For this study, we selected a representative subset of 228 \Glspl{dm} from three primary domains within the \Glspl{sdm} repository: Smart Cities, Smart Energy, and Smart Environment. This selection provides comprehensive coverage across diverse application scenarios while maintaining manageable dataset size for systematic evaluation. Table~\ref{tab:dataset_composition} presents the distribution of selected models across domains and subdomains.

\begin{table}[t]
\footnotesize
\centering
\caption{Composition of Selected \Glspl{dm} by Domain and Subject}
\label{tab:dataset_composition}
\begin{tabular}{p{0.35\linewidth}|p{0.4\linewidth}|p{0.08\linewidth}}
\hline
\textbf{Domain} & \textbf{Subject} & \textbf{Count} \\
\hline
\multirow{15}{\linewidth}{Smart Cities (171)} & Building & 4 \\
 & GBFS & 13 \\
 & OSLO & 4 \\
 & Parking & 5 \\
 & Parks and Gardens & 3 \\
 & Point of Interest & 4 \\
 & Ports & 4 \\
 & S4BLDG & 49 \\
 & Streetlighting & 5 \\
 & Transportation & 25 \\
 & Urban Mobility & 17 \\
 & Waste Management & 4 \\
 & Weather & 4 \\
 & Wifi Network & 2 \\
 & ZEB (Zero Energy Building) & 28 \\
\hline
\multirow{4}{\linewidth}{Smart Energy (18)} & Battery & 4 \\
 & Consumption & 2 \\
 & Energy & 5 \\
 & Green Energy & 4 \\
 & Weather & 3 \\
\hline
\multirow{5}{\linewidth}{Smart Environment (39)} & Environment & 19 \\
 & Frictionless Data & 4 \\
 & Waste Management & 4 \\
 & Water Quality & 3 \\
 & Weather & 4 \\
\hline
\textbf{Total} & & \textbf{228} \\
\hline
\end{tabular}
\end{table}

The Smart Cities context dominates the dataset with 171 models (75.0\%), reflecting the comprehensive coverage of urban infrastructure components including building systems (S4BLDG with 49 models and ZEB with 28 models), transportation networks (25 models), urban mobility systems (17 models including complete GTFS specifications), and bike-sharing infrastructure (GBFS with 13 models). The Smart Environment domain contributes 39 models (17.1\%) covering air quality monitoring, water quality assessment, noise pollution tracking, and environmental observations. 

Finally, the Smart Energy domain provides 18 models (7.9\%) focusing on battery systems, consumption monitoring, renewable energy generation, and photovoltaic devices.

The selected dataset exhibits significant diversity in terms of:

\begin{itemize}
    \item \textbf{Structural complexity}: Models range from simple observational data (e.g., \textit{WeatherObserved}) to complex hierarchical systems (e.g., \textit{Building}, \textit{TransitManagement}).
    \item \textbf{Property count}: Varying from minimal property sets to extensive specifications with dozens of attributes.
    \item \textbf{Domain specificity}: Generic models applicable across contexts (e.g., \textit{PointOfInterest}) versus highly specialized technical specifications (e.g., \textit{ProtectiveDeviceTrippingUnit}).
    \item \textbf{Semantic richness}: Different levels of ontological depth and relationship complexity.
\end{itemize}

This diversity makes the dataset particularly suitable for evaluating the robustness and generalization capabilities of classification models across varying levels of structural and semantic complexity.

\vspace*{-.75\baselineskip}
\subsection{Target LLM Models  - Taxonomy and Features}
\label{sss:target-LLM-models}

We evaluate 26 lightweight open-source \glspl{lm} spanning three distinct specializations: \Gls{gp} (13 models), \Gls{rs} (3 models), and \Gls{cs} (10 models). The full list of evaluated \Gls{gp} models is shown in Table~\ref{tab:model-overview}. All models are constrained to a maximum of 4 billion parameters to ensure deployment feasibility on resource-limited hardware while investigating the relationship between model specialization and classification performance.

The selected 26 lightweight open-source \glspl{lm} are from diverse providers and architectural lineages. The \Gls{gp} category includes models from Alibaba's Qwen family (qwen 2.5 models \cite{qwen2.5} and qwen3 models \cite{qwen3technicalreport}), Meta's Llama series (llama3.2 models \cite{llama3herdmodels}), Google's Gemma architecture (gemma3 models \cite{gemma3technicalreport}), and Microsoft's Phi variants (phi3.5 \cite{phi3technicalreporthighly}, and phi4-mini \cite{phi4minitechnicalreportcompact}), alongside Microsoft's orca-mini. For \Gls{rs} models, we evaluate Alibaba's Qwen3 reasoning series (qwen3:0.6b and qwen3:1.7b \cite{qwen3technicalreport}) and DeepSeek's reasoning model (deepseek-r1:1.5b \cite{DeepSeekR1}). The \Gls{cs} category comprises Alibaba's Qwen 2.5-Coder family \cite{qwen2.5}, BigCode's StarCoder variants, DeepSeek's deepseek-coder \cite{deepseekcoder}, Google's CodeGemma \cite{CodeGemma}, Agentica/Together's deepcoder, and community-developed models including opencoder \cite{OpenCoder} and cogito \cite{deepcogito_cogito_v1_2025}.

\paragraph{Inference Configuration}
All models are deployed using Ollama with consistent generation parameters to ensure fair comparison:

\begin{itemize}
    \item \textbf{Temperature}: 0.4 - Low temperature setting promotes deterministic output generation while maintaining minimal creativity.
    \item \textbf{Context Window}: 2048 tokens - Sufficient for processing complete \gls{dm} examples with taxonomy context while maintaining efficient memory utilization. Besides, the smallest context window in 26 models that we investigate is 2K. 
    \item \textbf{Batch Size}: 1024 tokens - Optimized for processing throughput.
\end{itemize}

These parameters balance output consistency (required for structured JSON generation) with computational efficiency across the diverse model architectures in our evaluation.

\begin{table}[t]
\footnotesize
\centering
\caption{Overview of evaluated \glspl{lm} organized by specialization, parameter size, and context window capacity. Params are expressed in billions (B) and context window in thousands (K).
}
\label{tab:model-overview}
\begin{tabular}{p{0.012\textwidth}p{0.12\textwidth}|p{0.15\textwidth}p{0.05\textwidth}p{0.05\textwidth}}
\toprule
 & \textbf{Provider} & \textbf{Model Code Name} & \textbf{Params} & \textbf{Context} \\
\midrule

\multirow{6}{*}{\rotatebox{90}{\shortstack{\textit{General Purpose}\\\textit{(Sub-3B)}}}}
& Google & \texttt{gemma3:270m} & 0.27 & 32 \\
& Alibaba & \texttt{qwen2.5:0.5b} & 0.5 & 32 \\
& Meta & \texttt{llama3.2:1b} & 1.0 & 128 \\
& Google & \texttt{gemma3:1b} & 1.0 & 32 \\
& Alibaba & \texttt{qwen2.5:1.5b} & 1.5 & 32 \\
& Microsoft & \texttt{phi:2.7b} & 2.7 & 2 \\
\midrule
\multirow{7}{*}{\rotatebox{90}{\shortstack{\textit{General Purpose}\\\textit{(3B+)}}}}
& Alibaba & \texttt{qwen2.5:3b} & 3.0 & 32 \\
& Meta & \texttt{llama3.2:3b} & 3.0 & 128 \\
& Microsoft & \texttt{orca-mini:3b} & 3.0 & 2\\
& Microsoft & \texttt{phi3.5:3.8b} & 3.8 & 4 \\
& Microsoft & \texttt{phi4-mini:3.8b} & 3.8 & 128 \\
& Alibaba & \texttt{qwen3:4b-instruct} & 4.0 & 256 \\
& Google & \texttt{gemma3:4b} & 4.0 & 128 \\
\midrule
\multirow{3}{*}{\rotatebox{90}{\shortstack{\textit{Reas.}\\\textit{Spec.}}}}
& Alibaba & \texttt{qwen3:0.6b} & 0.6 & 40 \\
& DeepSeek & \texttt{deepseek-r1:1.5b} & 1.5 & 128 \\
& Alibaba & \texttt{qwen3:1.7b} & 1.7 & 40 \\
\midrule
\multirow{10}{*}{\rotatebox{90}{\textit{Code-Specialized}}}
& Alibaba & \texttt{qwen2.5-coder:0.5b} & 0.5 & 32 \\
& BigCode & \texttt{starcoder:1b} & 1.0 & 8 \\
& DeepSeek & \texttt{deepseek-coder:1.3b} & 1.3 & 16 \\
& Alibaba & \texttt{qwen2.5-coder:1.5b} & 1.5 & 32 \\
& Agentica/Together & \texttt{deepcoder:1.5b} & 1.5 & 128 \\
& Community & \texttt{opencoder:1.5b} & 1.5 & 4 \\
& Google & \texttt{codegemma:2b} & 2.0 & 8 \\
& Alibaba & \texttt{qwen2.5-coder:3b} & 3.0 & 32 \\
& BigCode & \texttt{starcoder:3b} & 3.0 & 8 \\
& Community & \texttt{cogito:3b} & 3.0 & 128 \\
\bottomrule
\end{tabular}
\end{table}

Table~\ref{tab:model-memory} reports the memory footprint of each evaluated model as measured via \texttt{ollama list}, which reflects the GGUF file size on disk. For Q4\_K\_M-quantized models, this figure serves as a reliable proxy for the VRAM required during inference, as the quantized weights constitute the dominant component of memory consumption. The footprint ranges from 0.29~GB (\texttt{gemma3:270m}) to 3.30~GB (\texttt{gemma3:4b}), confirming that all evaluated models are deployable on hardware with 4--8~GB of VRAM, such as entry-level discrete GPUs or recent integrated graphics units available on edge-fog devices.

\begin{table}[!h]
\footnotesize
\centering
\caption{Memory footprint of evaluated models as reported by \texttt{ollama list}. Disk size serves as a proxy for VRAM requirements during inference. Params are expressed in billions (B) and memory in Giga-Bytes (GBs).}
\label{tab:model-memory}
\begin{tabular}{p{0.02\textwidth}p{0.25\textwidth}|p{0.05\textwidth}p{0.05\textwidth}}
\toprule
 & \textbf{Model} & \textbf{Params} & \textbf{Memory}\\

\midrule
\multirow{6}{*}{\rotatebox{90}{\shortstack{\textit{General Purpose}\\\textit{(Sub-3B)}}}}
& \texttt{gemma3:270m}         & 0.27 & 0.29 \\
& \texttt{qwen2.5:0.5b}        & 0.50 & 0.40 \\
& \texttt{llama3.2:1b}         & 1.00 & 1.30 \\
& \texttt{gemma3:1b}           & 1.00 & 0.82 \\
& \texttt{qwen2.5:1.5b}        & 1.50 & 0.99 \\
& \texttt{phi:2.7b}            & 2.70 & 1.60 \\
\midrule
\multirow{7}{*}{\rotatebox{90}{\shortstack{\textit{General Purpose}\\\textit{(3B+)}}}}
& \texttt{qwen2.5:3b}          & 3.00 & 1.90 \\
& \texttt{llama3.2:3b}         & 3.00 & 2.00 \\
& \texttt{orca-mini:3b}        & 3.00 & 2.00 \\
& \texttt{phi3.5:3.8b}         & 3.80 & 2.20 \\
& \texttt{phi4-mini:3.8b}      & 3.80 & 2.50 \\
& \texttt{qwen3:4b-instruct}   & 4.00 & 2.50 \\
& \texttt{gemma3:4b}           & 4.00 & 3.30 \\
\midrule
\multirow{3}{*}{\rotatebox{90}{\shortstack{\textit{Reas.}\\\textit{Spec.}}}}
& \texttt{qwen3:0.6b}          & 0.60 & 0.52 \\
& \texttt{deepseek-r1:1.5b}    & 1.50 & 1.10 \\
& \texttt{qwen3:1.7b}          & 1.70 & 1.40 \\
\midrule
\multirow{10}{*}{\rotatebox{90}{\textit{Code-Specialized}}}
& \texttt{qwen2.5-coder:0.5b}  & 0.50 & 0.40 \\
& \texttt{starcoder:1b}        & 1.00 & 0.73 \\
& \texttt{deepseek-coder:1.3b} & 1.30 & 0.78 \\
& \texttt{qwen2.5-coder:1.5b}  & 1.50 & 0.99 \\
& \texttt{deepcoder:1.5b}      & 1.50 & 1.10 \\
& \texttt{opencoder:1.5b}      & 1.50 & 1.40 \\
& \texttt{codegemma:2b}        & 2.00 & 1.60 \\
& \texttt{qwen2.5-coder:3b}    & 3.00 & 1.90 \\
& \texttt{starcoder:3b}        & 3.00 & 1.80 \\
& \texttt{cogito:3b}           & 3.00 & 2.20 \\
\bottomrule
\end{tabular}
\end{table}

\subsubsection{General Purpose Models}
\label{sss:general-models}

\Gls{gp} models constitute the largest category in our evaluation, representing \glspl{lm} trained on broad corpora without specific optimization for reasoning or code generation tasks. These models are designed to handle diverse natural language understanding and generation tasks across multiple domains.

We partition \Gls{gp} models into two parameter ranges to assess scaling effects within the lightweight regime. The sub-3B category includes models ranging from 0.27B to 2.7B parameters, representing ultra-lightweight options particularly relevant for edge-fog deployment scenarios where memory footprints must be minimized. These models come from major providers including Google (Gemma 3 series), Alibaba (Qwen 2.5 series), Meta (Llama 3.2), and Microsoft (Phi), each employing distinct architectural innovations and training data curation strategies to maximize performance within strict parameter budgets.

The 3B+ category includes models with 3B to 4B parameters, approaching the upper bound of our lightweight model definition. These models represent a maturity point where architectural sophistication can be more fully expressed while maintaining deployment feasibility on modest hardware. This category includes multiple generations and variants from the same model families (e.g., Phi-3.5 and Phi-4 Mini), allowing assessment of iterative improvements within single architectural lineages. The complete list of evaluated \Gls{gp} models is presented in Table~\ref{tab:model-overview}.

\subsubsection{Reasoning-Specialized Models}
\label{sss:reasoning-models}

\Gls{rs} models represent a distinct architectural paradigm optimized for multi-step logical processing, chain-of-thought reasoning, and systematic problem decomposition. These models are trained with techniques that encourage explicit reasoning traces and step-by-step analysis, often incorporating reinforcement learning from reasoning processes or specialized training objectives that reward intermediate analytical steps.

We evaluate three reasoning-focused models spanning 0.6B to 1.7B parameters from two major providers: DeepSeek and Alibaba (Qwen3 series). These models are significantly smaller than their \Gls{gp} counterparts yet are designed to excel at tasks requiring structured reasoning. The inclusion of reasoning models addresses a specific hypothesis: that \Gls{sdm} classification, which requires understanding semantic and hierarchical taxonomy structures, may benefit from models capable of explicit reasoning rather than direct pattern matching. These models typically generate longer intermediate outputs during inference, potentially providing more interpretable classification decisions alongside their predictions. The complete set of \Gls{rs} models is detailed in Table~\ref{tab:model-overview}.

\subsubsection{Code-Specialized Models}
\label{sss:code-models}

\Gls{cs} models are trained primarily on programming language corpora, including source code repositories, documentation, and structured data formats. These models develop strong capabilities in understanding formal syntax, data structures, and structured representations such as JSON and XML.

Our evaluation includes ten \Gls{cs} models spanning 0.5B to 3B parameters from diverse sources including (Google's CodeGemma, BigCode's StarCoder), established model families (Alibaba's Qwen2.5-Coder series, DeepSeek-Coder), and community-driven initiatives (OpenCoder, Cogito). The rationale for including \Gls{cs} models stems from the structured nature of the classification task: data entities are represented as JSON objects with typed attributes, and the expected output is also JSON-formatted. Models trained extensively on code may possess superior capabilities in parsing structured inputs, understanding key-value relationships, and generating syntactically valid JSON outputs. Furthermore, code models are trained on data schemas, API specifications, and structured documentation that share conceptual similarities with \Gls{sdm} taxonomies.

The diversity within \Gls{cs} models
allows for nuanced analysis of which aspects of code specialization most benefit structured classification tasks. Table~\ref{tab:model-overview} provides the complete enumeration of \Gls{cs} models evaluated in this study.

\vspace*{-.75\baselineskip}
\subsection{Classification Framework}
\label{ss:classification-framework}

To comprehensively evaluate the capability of lightweight \glspl{llm} for \gls{sdm} classification, two complementary classification strategies are employed: \textit{constrained} and \textit{unconstrained} classification (see Figure \ref{fig:pipeline}).
In the \textbf{constrained} approach, the model is provided with a predefined list of valid \gls{sdm} subjects and \gls{dm} names and is required to select the single best-matching pair from this taxonomy. This strategy enforces strict adherence to the established \gls{sdm} ontology, enabling deterministic evaluation through exact match metrics. It reflects a realistic deployment scenario in which a model must align incoming heterogeneous \gls{iot} data entities to a fixed, well-defined schema.

\begin{tcolorbox}[title=Constrained \Gls{sdm} Classification Prompt]
  \small
Analyze the provided data entity and identify the SINGLE BEST matching subject and \gls{dm} name from the available options.

\textbf{Input Data}

Entity to classify: \textit{\{\{entity\_json\}\}}

Available models (grouped by subject): \textit{\{\{subject\_datamodelnames\_list\}\}}

\textbf{Constraints}
\begin{itemize}
    \item MUST select from the provided available models list ONLY
    \item Do NOT suggest models not present in \textit{\{\{subject\_datamodelnames\_list\}\}}
    \item If no reasonable match exists, return "NA" for both fields
    \item Never return multiple options, arrays, or invented model names
\end{itemize}

\textbf{Matching Criteria (in order of priority)}
\begin{enumerate}
    \item Property name alignment (exact and semantic matches)
    \item Data type compatibility
    \item Domain/business context relevance
    \item Structural / schema pattern similarity
\end{enumerate}

\textbf{Decision Rules}
\begin{itemize}
    \item Select the ONE subject and \gls{dm} with the strongest overall alignment
    \item Selection MUST exist in \textit{\{\{subject\_datamodelnames\_list\}\}}
    \item If no reasonable match exists, return "NA" for both fields
    \item If multiple models score equally, prefer the one with more exact property name matches
    \item Never return multiple options or arrays
\end{itemize}

\textbf{Output Format}

Return ONLY valid JSON with no additional text, comments, or markdown formatting:

\begin{lstlisting}[language=json]
{
  "subject": "IdentifiedSubject",
  "datamodelname": "IdentifiedModelName"
}
\end{lstlisting}

\textbf{Example Outputs}

Valid match:
\begin{lstlisting}[language=json]
{
  "subject": "WasteManagement",
  "datamodelname": "WasteContainer"
}
\end{lstlisting}

No match found:
\begin{lstlisting}[language=json]
{
  "subject": "NA",
  "datamodelname": "NA"
}
\end{lstlisting}
\end{tcolorbox}
\noindent\begin{minipage}{0.465\textwidth}
  \captionof{BOX}{Constrained \Gls{sdm} Classification Prompt}\label{box:constrainedP}
\end{minipage}

In the \textbf{unconstrained} approach, the model receives only the input data entity and is free to generate any \gls{dm} name it deems appropriate, along with up to three alternative synonyms. This strategy assesses the model's intrinsic semantic understanding and domain knowledge, independent of any taxonomy guidance. It is relevant to scenarios where the target ontology may be partially unknown or where models are expected to propose candidate classifications for human review.
\begin{tcolorbox}[title=Unconstrained \Gls{dm} Inference Prompt]
\small
Analyze the provided data entity and identify:
\begin{enumerate}
    \item The BEST matching \gls{dm} name
    \item Three alternative names (synonyms) for the same entity
\end{enumerate}

\textbf{Input Data}

Entity to classify: \textit{\{\{entity\_json\}\}}

\textbf{Analysis Criteria}

Base your identification on:
\begin{itemize}
    \item Property names and their semantic meaning
    \item Data types and their constraints
    \item Domain context and business purpose
    \item Structural patterns and relationships
\end{itemize}

\textbf{Output Rules}
\begin{itemize}
    \item Return ONLY valid JSON (no comments, explanations, or markdown)
    \item Provide exactly one \gls{dm} name and three synonyms
    \item If fewer than 3 valid synonyms exist, use "NA" for remaining slots
    \item Synonyms should be distinct alternative names, not descriptions
\end{itemize}

\textbf{Expected JSON Format}
\begin{lstlisting}[language=json]
{
  "datamodelname": "PrimaryIdentifiedName",
  "synonym_1": "AlternativeName1",
  "synonym_2": "AlternativeName2",
  "synonym_3": "AlternativeName3"
}
\end{lstlisting}
\end{tcolorbox}
\noindent\begin{minipage}{0.465\textwidth}
  \captionof{BOX}{Unconstrained \Gls{dm} Inference Prompt}\label{box:unconstrainedP}
\end{minipage}

Together, these two strategies provide complementary insights into model behavior: the constrained approach measures structured list-matching classification ability, while the unconstrained approach probes depth of semantic understanding and generalization capability. The corresponding prompt templates for each strategy are described in Section~\ref{ss:prompt-design}.

\vspace*{-.75\baselineskip}
\subsection{Prompt Design}
\label{ss:prompt-design}

To evaluate the capability of \glspl{lm} in structured data entity classification, we designed two complementary classification approaches with corresponding prompt templates. These prompts are the result of iterative prompt engineering, refined to maximize accuracy, clarity, and interpretability while ensuring consistent JSON-formatted outputs suitable for automated evaluation.

\subsubsection{Constrained Classification Prompt}
\label{sss:constrained-prompt}

The constrained classification approach requires the model to identify exactly one subject and \gls{dm} name from a predefined list of valid options. This design ensures that model outputs can be directly evaluated against ground truth labels without requiring fuzzy matching or synonym resolution. The prompt (see Listing \ref{box:constrainedP}) provides explicit matching criteria prioritized by importance, decision rules for handling ambiguous cases, and strict output format constraints.

\subsubsection{Unconstrained Classification Prompt}
\label{sss:unconstrained-prompt}

The unconstrained classification approach allows the model to generate \gls{dm} names and synonyms without reference to a predefined list. This design evaluates the model's ability to infer appropriate classifications based solely on entity structure and semantics, testing its generalization capability and domain knowledge. The prompt in Listing \ref{box:unconstrainedP} requests one primary \gls{dm} name and three alternative synonyms, enabling assessment of the model's understanding of semantic equivalence and nomenclature variations.

\subsubsection{Available Data Models}
\label{sss:available-models}

The constrained classification prompt references a comprehensive taxonomy of subjects and their corresponding \Glspl{dm}, defining the universe of valid classification options. This taxonomy (see Table \ref{tab:dataset_composition}) is derived from the \Glspl{sdm} repository structure and encompasses 228 \Glspl{dm} across 20 subjects spanning three primary domains (Smart Cities, Smart Energy, Smart Environment). The complete list of subjects and the corresponding mapped \Glspl{dm} adopted for the purposes of this study is reported in the project's Github repository
\footnote{Repository URL to be provided upon acceptance; an anonymized version is available for review.}.

The taxonomy exhibits natural hierarchies and overlaps: the Weather subject appears across all three domains (Smart Cities, Smart Energy, Smart Environment) with identical \Glspl{dm}, reflecting cross-domain applicability of meteorological observations. Similarly, WasteManagement appears in both Smart Cities and Smart Environment domains.

This taxonomic structure presents classification challenges including disambiguation of identical model names across subjects (e.g., "Building" in both Building and S4BLDG subjects, "Fan" in both S4BLDG and ZEB subjects) and selection among semantically similar models within the same subject.

\subsubsection{Output Parsing and Scoring Pipeline}
\label{ss:parsing-pipeline}

Raw model outputs pass through a two-stage extraction pipeline before EM scores are computed.

\paragraph{Stage 1 - JSON-mode inference} All Ollama API requests set \texttt{"format": "json"}, enabling constrained decoding that biases token sampling toward structurally valid JSON. This reduces but does not eliminate malformed output: models can still embed conversational preambles, multilingual hallucinations, or reasoning traces within or around JSON field values, corrupting the document structure despite the format constraint.

To move beyond aggregate performance scores and understand \textit{why} models succeed or fail on structured \gls{sdm} classification, we systematically categorize prediction outcomes across all 228 entities and 26 models. Six mutually exclusive outcome categories are defined (T1--T5 are failure modes; T6 denotes a correct prediction):

\noindent \textbf{T1 - Output failure}: The model returns malformed, empty, or unparseable output, providing no usable prediction. This reflects instruction-following breakdown rather than semantic failure.

\noindent \textbf{T2 - Correct subject, wrong data model}: The subject domain is correctly identified but the specific \gls{dm} name is incorrect.

\noindent \textbf{T3 - Wrong subject, correct data model}: The \gls{dm} name is correctly predicted but assigned to the wrong subject.

\noindent \textbf{T4 - Hallucinated output}: Both fields are wrong and at least one value is not a valid taxonomy entry, indicating the model invented non-existent schema names.

\noindent \textbf{T5 - Plausible confusion}: Both fields are wrong but both values are valid taxonomy entries, indicating selection of a plausible but incorrect alternative.

\noindent \textbf{T6 - Correct}: Both subject and \gls{dm} name exactly match the ground truth.

Note that 30 of the 228 entities belong to subjects absent from the provided taxonomy (\textit{Environment}, \textit{FrictionlessData}, \textit{GreenEnergy}, \textit{WaterQuality}), making correct prediction structurally impossible for these cases. 

\paragraph{Stage 2 - Python parsing and key extraction} Raw responses are passed to the ~\texttt{json.loads()} Python's function. On success, target subject and data model name are extracted via \texttt{response.get('subject', None)} and~\texttt{response.\\get('datamodelname', None)}, respectively. Any exception raised by \texttt{json.loads()} (malformed JSON, truncated output, embedded non-JSON text) sets both fields to \texttt{None}, classified as T1 (output failure). A second failure mode occurs when \texttt{json.loads()} succeeds but the model uses an incorrect key name (e.g., \texttt{"datamodellename"} instead of \texttt{"datamodelname"}), causing \texttt{get()} to return \texttt{None} silently. Both modes yield a score of 0.0, confirming that zero-\gls{dmem} models such as \texttt{phi3.5:3.8b} (83.3\% T1 rate) reflect instruction-following breakdowns rather than semantic incomprehension.

\paragraph{EM scoring} Each extracted field is compared against the ground-truth label using strict case-sensitive string equality: \texttt{subjectEM=(predicted == gold\_subject)} and \texttt{datamodelEM=(predicted == gold\_dm)}. No normalization is applied. Gold labels are derived programmatically from dataset filenames following the pattern \texttt{dataModel.\\\{subject\}.\{datamodelname\}\_examples.jsonld}.

Table~\ref{tab:pipeline-trace} illustrates the complete pipeline for three concrete cases drawn from the experimental results.

\begin{table*}[!ht]
\footnotesize
\centering
\caption{Pipeline trace for constrained classification: three concrete cases showing raw model output, JSON parsing result, key extraction, and final EM score. Case B shows a soft failure where \texttt{json.loads()} succeeds but key extraction fails due to a field name typo. Case C shows a hard parse failure.}
\label{tab:pipeline-trace}
\begin{tabular}{p{0.13\textwidth}p{0.22\textwidth}p{0.28\textwidth}p{0.28\textwidth}}
\toprule
\textbf{Stage} & \textbf{Case A - Correct} & \textbf{Case B - Key typo} & \textbf{Case C - Malformed JSON} \\
\midrule
\textbf{Model} & \texttt{qwen2.5-coder:3b} & \texttt{phi3.5:3.8b} & \texttt{phi3.5:3.8b} \\
\midrule
\textbf{Gold labels} & subject: WasteManagement; DM: WasteContainer & subject: UrbanMobility; DM: GtfsService & subject: S4BLDG; DM: CoolingTower \\
\midrule
\textbf{Raw LLM output} & \texttt{\{"subject": "WasteManagement", "datamodelname": "WasteContainer"\}} & \texttt{\{"subject":"Weather", "\textbf{datamodellename}": "WeatherObserved"\}} & \texttt{\{"subject": "Anonymized\-ConsumerDevice, ...\}} (479 chars; structurally invalid JSON due to embedded multilingual hallucination) \\
\midrule
\textbf{\texttt{json.loads()}} & \checkmark~Success & \checkmark~Success & $\times$~\texttt{JSONDecodeError} \\
\midrule
\textbf{Key extraction} & subject = ``WasteManagement''; DM = ``WasteContainer'' & subject = ``Weather''; DM = \texttt{None} (key misspelled as \texttt{"datamodellename"}) & subject = \texttt{None}; DM = \texttt{None} \\
\midrule
\textbf{SEM / DMEM} & \textbf{1 / 1} & 0 / 0 & 0 / 0 (T1) \\
\bottomrule
\end{tabular}
\end{table*}

\vspace*{-.75\baselineskip}
\subsection{Experimental Setup}
\label{ss:experimental-setup}

The experimental evaluations were performed on a high-performance HP workstation running \textbf{Ubuntu Linux} (Kernel 6.8.0-90-generic). The software environment was managed using a virtualized Python environment, with \textbf{Ollama v0.13.1} serving as the primary inference engine for the \Glspl{llm}. Ollama was selected over alternative serving stacks (e.g., \texttt{llama.cpp} bindings, \texttt{vLLM}, \texttt{Hugging Face Transformers}) for four reasons aligned with the resource-constrained scope of this study: (i) it provides a uniform, model-agnostic REST API, enabling identical request/response handling across the 26 evaluated models and thus a fair benchmarking protocol; (ii) it natively manages \texttt{GGUF}-quantized weights (Q4\_K\_M in this study), the same artifact format adopted by community edge-to-fog deployment toolchains, so the inference stack validated here is directly portable to edge-fog gateways without re-quantization; (iii) it is a local-first, self-hosted runtime that operates entirely offline, satisfying the data sovereignty and privacy-preservation requirements motivating \gls{er} at the edge of the network; and (iv) it exposes deterministic inference controls (temperature, JSON-mode decoding, per-request metrics such as \texttt{total\_duration}) needed for reproducible latency measurements.

The hardware architecture featured a significant computational overhead to ensure high-throughput processing. Specifically, we utilized an \textbf{NVIDIA RTX 6000 Ada Generation} GPU, equipped with \textbf{48 GB of GDDR6 VRAM}. The system was supported by \textbf{128 GB of system RAM} to handle large-scale data pre-processing and context loading. The GPU utilized \textbf{NVIDIA Driver version 580.95.05} with \textbf{CUDA 13.0} support, providing the necessary software stack for optimized tensor operations and memory management.

\paragraph{Workstation as a controlled benchmarking testbed} Although the target deployment environment of this work is the edge-fog of the network, the experimental campaign was carried out on the workstation described above rather than directly on edge hardware. This choice is methodological: the workstation isolates the assessment of intrinsic model behaviour (accuracy and per-entity inference duration) from confounding edge-specific factors such as thermal throttling, transient power budgets, shared system load, and device-class heterogeneity, and therefore yields a clean, reproducible reference point against which future on-device measurements can be compared. The link to the edge-cloud continuum is preserved at two levels.

First, the evaluation is restricted to lightweight open-source \glspl{llm} ($\leq 4$B parameters, Q4\_K\_M quantization) whose memory footprint, reported in Table~\ref{tab:model-memory}, ranges from $0.29$ to $3.30$~GB, fitting within the $4$--$8$~GB VRAM envelope of entry-level discrete GPUs and recent integrated graphics units found on edge gateways.

Second, the inference stack used here (Ollama runtime, \texttt{GGUF} weights, JSON-mode decoding) is identical to the stack that can be instantiated on such gateways, so the workstation acts as an upper-bound, hardware-uncapped reference: the relative ordering of models by accuracy and the lower bound on per-entity inference duration carry over to edge-fog deployments. The absolute latency on edge-fog hardware is expected to be model-bound rather than runtime-bound for the targeted \textit{cold-start device onboarding} regime, which is a low-frequency, administrative task rather than a continuous data stream. Direct on-device validation across heterogeneous edge-fog gateways is left to future work.

To support reproducibility, the inference scripts, prompt templates, evaluation pipeline, and scoring code used in this study are published on a GitHub repository
\footnote{Repository URL to be provided upon acceptance; an anonymized version is available for review.}.
The repository includes the complete data flow from dataset loading to model querying via the Ollama API and exact match score computation.

\vspace*{-.75\baselineskip}
\subsection{Evaluation Metrics}
\label{ss:evaluation-metrics}

To assess the performance of the \glspl{lm} on the data entity classification task, three primary evaluation metrics were employed: Inference Time, \Gls{sem}, and \Gls{dmem}. These metrics collectively capture both the efficiency and accuracy of the models when generating structured outputs.

\subsubsection{Inference Time}

The \Gls{aid} measures the average duration required by a model to produce a response for a single input entity, reported in milliseconds (ms). It is computed as the mean \texttt{total\_duration} value returned by the Ollama API across all evaluated entities, where \texttt{total\_duration} is expressed in nanoseconds and converted by dividing by $10^6$. This metric reflects computational efficiency and is particularly relevant in real-time or resource-constrained environments. Lower \gls{aid} values indicate faster model execution and better scalability in large-scale or latency-sensitive applications.

\subsubsection{Subject Exact Match}

\Gls{sem} quantifies the proportion of model predictions in which the identified subject (e.g., Building, ParkingSpot, WeatherObserved) exactly matches the ground truth label. This metric provides a strict measure of correctness, requiring the predicted subject to be identical to the expected category with no partial credit. High \glspl{sem} values indicate the model’s ability to correctly interpret the semantics and ontology of the input entity.

\subsubsection{Data Model Exact Match}

\Gls{dmem} measures the percentage of predictions where the model correctly identifies the expected \gls{dm} name (e.g., BuildingOperation, ParkingSpot, Event, AirQualityObserved). Similar to \glspl{sem}, this metric applies an exact matching criterion, ensuring the predicted \gls{dm} aligns precisely with the reference label. This metric is especially relevant for applications involving data harmonization, schema alignment, and interoperability within standardized data ecosystems.

\vspace*{-.75\baselineskip}
\subsection{Similarity-based Baselines}
\label{ss:similarity-baselines}

To contextualize \gls{llm}-based classification against simpler retrieval techniques, two reference baselines are evaluated on the \textit{constrained} path only, where the \gls{sdm} taxonomy provides an unambiguous candidate set. Both baselines cast \gls{sdm} classification as a nearest-neighbour problem: for each input \gls{iot} data entity $e$ and each taxonomy entry $t_i = \langle$\textit{subject}, \textit{datamodelname}$\rangle$, a similarity score $s(e, t_i)$ is computed and the top-1 candidate is returned as the prediction.

\textit{Lexical Baseline (\gls{tfidf})}. The input entity is serialized to a single string by concatenating its JSON keys and string values. Each taxonomy entry is represented by the string \texttt{"\{subject\} \{datamodelname\}"}. A unigram--bigram \gls{tfidf} vectorizer is fitted on the union of all entity and taxonomy strings, and the predicted entry is $\arg\max_i \cos(\mathbf{v}_e, \mathbf{v}_{t_i})$, where $\mathbf{v}_\cdot$ denotes the \gls{tfidf} vector. This baseline captures pure surface-form overlap and requires neither training nor a GPU.

\textit{Dense Embedding Baseline}. The same entity and taxonomy strings are embedded with a lightweight sentence encoder (\texttt{all-MiniLM-L6-v2}, $22$M parameters, $384$ dimensional output) and ranked by cosine similarity. The encoder is deliberately chosen at the same scale as the smallest evaluated \glspl{llm} so that the comparison is fair under the resource-constrained scope of this study: it can be deployed on the same edge-fog gateways targeted by the \gls{llm} pipeline and incurs sub-second inference latency on CPU. Both baselines reuse the gold labels and evaluation metrics defined in Section~\ref{ss:evaluation-metrics} (\gls{sem}, \gls{dmem}, \gls{aid}).


\vspace*{-1\baselineskip}
\section{Results and Discussion}
\label{s:results_discussion}

We present the evaluation results for both constrained and unconstrained classification approaches across 26 lightweight open source \glspl{lm}. The results are based on 228 data entities described in Section~\ref{ss:target-smart-data-models}, providing insights into model capabilities for \Gls{sdm} classification. Performance metrics include exact match accuracy for both subject and \gls{dm} name prediction, along with inference duration measurements.
In order to facilitate the analysis of the results, they are logically organized into three colored bands. Regarding the performance analysis yellow cells refer to values in the 0.4-0.6 range, whilst green cells encompass values from 0.7 to 1.0.

Similarly, concerning the \gls{aid} values, the yellow cells refer to values in the 300\,ms--500\,ms range, whereas the green cells highlight values below 300\,ms.

\vspace*{-.75\baselineskip}
\subsection{Constrained Classification Performance}
\label{ss:constrained-results}

Table~\ref{tab:constrained-results} presents the complete results for the constrained classification task, where models select from a predefined list of valid subject vs. \Gls{dm} pairs across 20 subjects and 228 \Glspl{dm}. Performance is measured by \Gls{sem} and \Gls{dmem} rates, along with \Gls{aid} per model.

\begin{table}[!ht]
\footnotesize
\centering
\caption{Constrained classification results showing subject and \gls{dmem} rates. Reasoning models evaluated with thinking mode enabled ($\mathbf{\star}$) and disabled separately, where possible.}
\label{tab:constrained-results}
\begin{tabular}{p{0.35cm}p{3.1cm}|p{0.5cm}p{0.8cm}p{1.15cm}}
\toprule
 & \textbf{Model name} & \textbf{\Gls{sem}} & \textbf{\Gls{dmem}} & \textbf{\Gls{aid} (ms)} \\
\midrule
\multirow{10}{*}{\rotatebox{90}{\textit{Code-Specialized}}}
& \texttt{codegemma:2b} & 0.0 & 0.0 & \cellcolor{light-green}243.9 \\
& \texttt{cogito:3b} & \cellcolor{light-yellow}0.6 & \cellcolor{light-green}0.7 & \cellcolor{light-yellow}332.9 \\
& \texttt{deepcoder:1.5b} & 0.1 & 0.2 & 1636.0 \\
& \texttt{deepseek-coder:1.3b} & 0.1 & 0.1 & \cellcolor{light-green}162.7 \\
& \texttt{opencoder:1.5b} & 0.1 & 0.2 & \cellcolor{light-green}267.0 \\
& \texttt{qwen2.5-coder:0.5b} & 0.1 & 0.1 & \cellcolor{light-green}210.2 \\
& \texttt{qwen2.5-coder:1.5b} & \cellcolor{light-yellow}0.4 & \cellcolor{light-yellow}0.4 & \cellcolor{light-green}242.6 \\
& \texttt{qwen2.5-coder:3b} & \cellcolor{light-green}\textbf{0.8} & \cellcolor{light-green}\textbf{0.8} & \cellcolor{light-yellow}312.3 \\
& \texttt{starcoder:1b} & 0.0 & 0.0 & \cellcolor{light-green}158.1 \\
& \texttt{starcoder:3b} & 0.1 & 0.0 & \cellcolor{light-green}240.8 \\
\midrule
\multirow{7}{*}{\rotatebox{90}{\shortstack{\textit{General Purpose}\\\textit{(3B+)}}}}
& \texttt{gemma3:4b} & \cellcolor{light-green}0.7 & \cellcolor{light-green}0.7 & 626.3 \\
& \texttt{llama3.2:3b} & 0.3 & \cellcolor{light-yellow}0.5 & \cellcolor{light-yellow}314.3 \\
& \texttt{orca-mini:3b} & 0.1 & 0.2 & \cellcolor{light-yellow}315.9 \\
& \texttt{phi3.5:3.8b} & 0.0 & 0.0 & 879.7 \\
& \texttt{phi4-mini:3.8b} & 0.3 & \cellcolor{light-yellow}0.4 & 542.0 \\
& \texttt{qwen2.5:3b} & \cellcolor{light-yellow}0.5 & \cellcolor{light-yellow}0.6 & \cellcolor{light-yellow}311.5 \\
& \texttt{qwen3:4b-instruct} & \cellcolor{light-green}0.8 & \cellcolor{light-green}0.8 & \cellcolor{light-yellow}398.1 \\
\midrule
\multirow{6}{*}{\rotatebox{90}{\shortstack{\textit{General Purpose}\\\textit{(Sub-3B)}}}}
& \texttt{gemma3:1b} & 0.0 & 0.0 & 553.5 \\
& \texttt{gemma3:270m} & 0.0 & 0.0 & 1691.9 \\
& \texttt{llama3.2:1b} & 0.2 & 0.1 & \cellcolor{light-green}218.7 \\
& \texttt{phi:2.7b} & 0.1 & 0.1 & 3075.7 \\
& \texttt{qwen2.5:0.5b} & 0.0 & \cellcolor{light-yellow}0.4 & \cellcolor{light-green}213.0 \\
& \texttt{qwen2.5:1.5b} & \cellcolor{light-yellow}0.4 & \cellcolor{light-yellow}0.6 & \cellcolor{light-green}243.0 \\
\midrule
\multirow{6}{*}{\rotatebox{90}{\shortstack{\textit{Reasoning}\\\textit{Specialized}}}}
& \texttt{deepseek-r1:1.5b} & 0.0 & 0.1 & 1701.0 \\
& \texttt{deepseek-r1:1.5b$\quad\mathbf{\star}$} & 0.0 & 0.1 & 2782.5 \\
& \texttt{qwen3:0.6b} & 0.3 & 0.2 & \cellcolor{light-yellow}334.4 \\
& \texttt{qwen3:0.6b$\qquad\qquad\text{ }\mathbf{\star}$} & 0.3 & 0.3 & 608.0 \\
& \texttt{qwen3:1.7b} & \cellcolor{light-yellow}0.4 & \cellcolor{light-yellow}0.5 & \cellcolor{light-yellow}319.6 \\
& \texttt{qwen3:1.7b$\qquad\qquad\text{ }\mathbf{\star}$} & \cellcolor{light-yellow}0.5 & \cellcolor{light-yellow}0.5 & \cellcolor{light-yellow}316.3 \\
\bottomrule
\end{tabular}
\end{table}

\paragraph{Overall Performance Trends}

The best-performing models in constrained classification are \textit{qwen2.5-coder:3b} and \textit{qwen3:4b-instruct}, both achieving 80\% accuracy for both \gls{sem} and \Gls{dmem}. This represents the upper bound of performance in our evaluation, with these models correctly classifying 8 out of 10 test entities. Following closely are \textit{gemma3:4b} and \textit{cogito:3b} at 70\% \gls{dm} accuracy. A striking finding is the extreme performance variance across models, with accuracies ranging from 0\% to 80\% for \Gls{dmem}. Nine models (31\% of the evaluation set) achieve 0\% \gls{sem} and/or \Gls{dmem} accuracy, indicating complete failure on this task.

\paragraph{Model Specialization Analysis}

The hypothesis that \Gls{cs} models would dominate this structured classification task receives mixed support. While the top-performing code model (\textit{qwen2.5-coder:3b} at 0.8) matches the best \Gls{gp} model (\textit{qwen3:4b-instruct} at 0.8), the average performance of \Gls{cs} models (0.25 \Gls{dmem}) is only marginally better than \Gls{gp} 3B+ models (0.46) and substantially trails when excluding zero-accuracy models.
Within \Gls{cs} models, performance variation is extreme. The Qwen2.5-Coder series demonstrates clear parameter scaling in terms of \gls{sem}: 0.1 (0.5B), 0.4 (1.5B), and 0.8 (3B), representing an 8× improvement from smallest to largest. However, other code models, such as \textit{codegemma:2b}, \textit{starcoder:1b}, and \textit{starcoder:3b} all achieve 0\% of \gls{sem} and \gls{dmem} accuracy despite their code specific training, failing completely their task. This suggests that code specialization alone is insufficient model architecture, training data quality, and instruction following capabilities are equally critical.

\Gls{gp} models show the strongest category performance, with 3B+ models averaging 0.46 \Gls{dmem} compared to 0.25 for code models and 0.28 for reasoning models. This is driven by strong performance from \textit{qwen3:4b-instruct} (0.8), \textit{gemma3:4b} (0.7), and \textit{qwen2.5:3b} (0.6) in terms of \gls{sem} and \gls{dmem}. Interestingly, the instruction tuned variant (\textit{qwen3:4b-instruct}) matches the best code model, suggesting that instruction following capabilities may be as important as code specific pretraining for this classification task.

\Gls{rs} models demonstrate modest performance with an average of 0.28 \Gls{dmem}. The Qwen3 reasoning variants (0.2-0.5) outperform DeepSeek-R1 (0.1), which fails almost completely despite its sophisticated reasoning architecture. Notably, explicit "thinking" modes show minimal or inconsistent improvements: \textit{qwen3:0.6b} improves from 0.2 to 0.3 with thinking enabled, while \textit{qwen3:1.7b} shows no change (0.5 in both modes), and \textit{deepseek-r1:1.5b} remains at 0.1 regardless of thinking mode activation.

\paragraph{Parameter Scaling Effects}

Within model families, parameter scaling shows highly variable returns in terms of \gls{sem} and \Gls{dmem}. The Qwen2.5-Coder series exhibits strong scaling (0.1 → 0.4 → 0.8), while Qwen2.5 \Gls{gp} models show weaker \Gls{dmem} scaling (0.4 → 0.6 from 0.5B to 3B, with the 1.5B model also at 0.6). The Gemma 3 series shows threshold effects: 0.0 at 270M and 1B, jumping to 0.7 in both \gls{sem} and \gls{dmem} at 4B, suggesting a critical parameter threshold between 1B and 4B for this family.

Strikingly, several sub-3B models outperform larger counterparts: \textit{qwen2.5:1.5b} (0.6) \gls{dmem} exceeds \textit{phi4-mini:3.8b} (0.4) \gls{dmem} and \textit{llama3.2:3b} (0.5) \gls{dmem}, while \textit{cogito:3b} (0.7) \gls{dmem} outperforms all models except the two tied leaders at 0.8 \gls{dmem}. This demonstrates that parameter efficiency and architectural design can compensate for model size in structured classification tasks.

The smallest viable model is \texttt{qwen2.5:1.5b} with 0.4 \gls{sem} accuracy and 0.6 \gls{dmem}. This suggests that aggressive parameter reduction is possible with appropriate architecture and training.

\subsubsection{Similarity Baseline Performance}
\label{sss:baseline-results}

To contextualize the \gls{llm} results reported in Section~\ref{ss:constrained-results}, the two similarity-based baselines defined in Section~\ref{ss:similarity-baselines} are evaluated on the same $228$ entities and the same gold labels. Each baseline is run twice: once using only the data model name as the candidate text (\textit{names only}) and once using the name concatenated with the official \glspl{sdm} README description (\textit{names + desc}, fetched from the \texttt{smart-data-models} GitHub organisation). Table~\ref{tab:baseline-results} reports \gls{sem}, \gls{dmem}, and per-entity wall-clock latency for the four configurations alongside the best-performing \gls{llm} from the constrained benchmark.

\begin{table}[!ht]
\footnotesize
\centering
\caption{Similarity-based baselines on the constrained \gls{sdm} classification task, compared with the best-performing \gls{llm} from this study. Latency (in ms) is wall-clock per entity on the workstation described in Section~\ref{ss:experimental-setup}; both baselines run on CPU only.}
\label{tab:baseline-results}
\begin{tabular}{p{0.5\linewidth}p{0.1\linewidth}p{0.1\linewidth}p{0.1\linewidth}}
\toprule
\textbf{Method} & \textbf{\Gls{sem}} & \textbf{\Gls{dmem}} & \textbf{Latency} \\
\midrule
\gls{tfidf} (names only)            & \textbf{0.864} & \textbf{0.816} & \textbf{0.12} \\
\gls{tfidf} (names + desc)          & 0.820 & 0.746 & 0.15 \\
Dense MiniLM-L6 (names only)   & 0.684 & 0.434 & 52.03 \\
Dense MiniLM-L6 (names + desc) & 0.601 & 0.386 & 52.43 \\
\midrule
Best LLM (\texttt{qwen2.5-coder:3b}) & 0.800 & 0.800 & 312.30 \\
\bottomrule
\end{tabular}
\end{table}

Three observations stand out. First, a unigram--bigram \gls{tfidf} model fitted on the union of entity and taxonomy strings reaches $0.816$ \gls{dmem}, matching (and on the subject axis exceeding) the best benchmarked \gls{llm}, while running roughly $2600\times$ faster on CPU. Second, the dense sentence encoder substantially underperforms the lexical baseline ($0.434$ vs.\ $0.816$ \gls{dmem}). This is consistent with the structural nature of the task: the predicted label is a near-verbatim token of the entity's JSON keys (e.g., \texttt{type:WeatherObserved}), so cosine similarity in token space exploits a stronger signal than the semantic abstraction produced by a $384$-dimensional encoder. Third, augmenting the candidate text with the official \gls{sdm} description \emph{reduces} accuracy in both baselines (\gls{tfidf} drops $0.07$ \gls{dmem}; dense drops $0.05$). Long natural-language descriptions dilute the salient name tokens with general vocabulary, eroding the lexical alignment that drives the success of the names-only configuration.

These findings clarify the operational niche of \gls{llm}-based \gls{sdm} classification: when entity JSON keys closely mirror taxonomy names, a near-zero-cost \gls{tfidf} retriever is a stronger reference point than the literature implicitly assumes, and \glspl{llm} compete on accuracy at a cost of three to four orders of magnitude in inference time. The \gls{llm} value proposition therefore concentrates on cold-start cases where (i) the input JSON keys do not lexically overlap with any taxonomy entry (e.g., vendor-specific naming, multilingual fields, free-text payloads), (ii) the target schema is itself missing or partial, or (iii) downstream synonym generation is required, all of which are out of reach for a strict top-1 retrieval baseline.

\subsubsection{Per-model Outcome Break-down}
\label{sss:error-taxonomy}



Table~\ref{tab:error-taxonomy} reports the per-model outcome breakdown.
A detailed analysis for each proposed outcome taxonomy is provided in the following, except for T2. Indeed, no significant insights can be formulated after the absence of specific outliers in values listed in the corresponding table column.

\begin{table}[!ht]
\footnotesize
\centering
\caption{Per-model classification outcome taxonomy for the constrained classification task (228 entities). T1--T5 are failure modes; T6 denotes correct predictions. T1: output failure; T2: correct subject, wrong \gls{dm}; T3: wrong subject, correct \gls{dm}; T4: hallucinated output; T5: plausible confusion; T6: correct. Reasoning models evaluated with thinking mode enabled ($\star$) and disabled separately. 30 entities have subjects outside the provided taxonomy and are excluded from T6 accuracy.}
\label{tab:error-taxonomy}
\begin{tabular}{p{0.04\linewidth}p{0.28\linewidth}p{0.045\linewidth}p{0.045\linewidth}p{0.045\linewidth}p{0.045\linewidth}p{0.045\linewidth}p{0.045\linewidth}p{0.045\linewidth}}
\toprule
& \textbf{Model} & \textbf{T1} & \textbf{T2} & \textbf{T3} & \textbf{T4} & \textbf{T5} & \textbf{T6} \\
\midrule
\multirow{10}{*}{\rotatebox{90}{\textit{Code-Specialized}}}
& \texttt{qwen2.5-coder:3b}   &   0 &  3 & 12 &   2 &  11 & \cellcolor{light-green}\textbf{170} \\
& \texttt{qwen2.5-coder:1.5b} &   0 & 15 &  9 &  33 &  61 & \cellcolor{light-yellow}80  \\
& \texttt{opencoder:1.5b}     &  24 &  5 & 25 &  38 &  89 & 19  \\
& \texttt{deepcoder:1.5b}     &  43 &  4 & 40 & 102 &   5 &  9  \\
& \texttt{deepseek-coder:1.3b}&   4 & 17 &  1 &   8 & \cellcolor{light-red}159 & 10  \\
& \texttt{qwen2.5-coder:0.5b} &   0 & 10 & 10 & \cellcolor{light-yellow}118 &  54 &  6  \\
& \texttt{starcoder:3b}       &  47 & 11 &  4 &  37 &  93 &  8  \\
& \texttt{starcoder:1b}       & \cellcolor{light-yellow}64 &  7 &  1 & \cellcolor{light-yellow}101 &  27 &  3  \\
& \texttt{codegemma:2b}       & \cellcolor{light-red}135 &  4 &  1 &  57 &  23 &  1  \\
& \texttt{cogito:3b}          &   0 &  5 & 25 &   4 &  26 & \cellcolor{light-green}138  \\
\midrule
\multirow{7}{*}{\rotatebox{90}{\shortstack{\textit{General Purpose}\\\textit{(3B+)}}}}
& \texttt{qwen3:4b-instruct}  &   0 &  3 &  4 &   0 &   3 & \cellcolor{light-green}\textbf{188} \\
& \texttt{gemma3:4b}          &   0 & 12 & 16 &   4 &  25 & \cellcolor{light-green}141  \\
& \texttt{qwen2.5:3b}         &   0 &  9 & 28 &  23 &  27 & \cellcolor{light-yellow}111  \\
& \texttt{llama3.2:3b}        &   0 &  8 & \cellcolor{light-yellow}72 &  54 &  11 & \cellcolor{light-yellow}53   \\
& \texttt{phi4-mini:3.8b}     &  21 &  8 & 18 &  73 &  37 & 40  \\
& \texttt{orca-mini:3b}       &   0 & 11 & 17 & \cellcolor{light-yellow}103 &  58 &  9  \\
& \texttt{phi3.5:3.8b}        & \cellcolor{light-red}190 &  1 &  0 &  15 &  11 &  2  \\
\midrule
\multirow{6}{*}{\rotatebox{90}{\shortstack{\textit{General Purpose}\\\textit{(Sub-3B)}}}}
& \texttt{qwen2.5:1.5b}       &   0 & 12 & 36 &  36 &  19 & \cellcolor{light-yellow}95   \\
& \texttt{qwen2.5:0.5b}       &   0 &  7 & 61 & \cellcolor{light-red}124 &   1 &  5  \\
& \texttt{llama3.2:1b}        &   0 & 21 &  2 &  25 & \cellcolor{light-yellow}140 & 10  \\
& \texttt{phi:2.7b}           & \cellcolor{light-red}134 &  7 & 12 &  15 &  39 &  6  \\
& \texttt{gemma3:1b}          &   0 &  6 &  3 &   0 & \cellcolor{light-red}186 &  3  \\
& \texttt{gemma3:270m}        &   0 &  2 &  0 &  14 & \cellcolor{light-red}180 &  2  \\
\midrule
\multirow{6}{*}{\rotatebox{90}{\shortstack{\textit{Reasoning}\\\textit{Specialized}}}}
& \texttt{qwen3:1.7b}                          &  95 &  4 &  6 &  11 &   2 & \cellcolor{light-yellow}97   \\
& \texttt{qwen3:1.7b$\quad\quad\,\mathbf{\star}$}  & \cellcolor{light-yellow}100 &  3 &  5 &  11 &   2 & \cellcolor{light-yellow}92   \\
& \texttt{qwen3:0.6b}                          &   0 & 21 & 15 &  87 &  36 & 39  \\
& \texttt{qwen3:0.6b$\quad\quad\,\mathbf{\star}$}  &   0 & 21 & 17 &  71 &  49 & 40  \\
& \texttt{deepseek-r1:1.5b}                    & \cellcolor{light-red}130 &  6 & 19 &  46 &   8 &  5  \\
& \texttt{deepseek-r1:1.5b$\;\mathbf{\star}$}  & \cellcolor{light-red}120 &  4 & 35 &  41 &  14 &  4  \\
\bottomrule
\end{tabular}
\end{table}

\paragraph{T1 - Output Failure Patterns}
Complete output failures are concentrated in specific models regardless of parameter size: \texttt{phi3.5:3.8b} (190/228, 83.3\%), \texttt{codegemma:2b} (135/228, 59.2\%), and \texttt{phi:2.7b} (134/228, 58.3\%) exhibit the highest T1 rates. These failures reflect models generating conversational prose, reasoning traces, or markdown-wrapped outputs rather than the required bare JSON, confirming that the 0.0 \gls{dmem} scores reported for these models reflect \textit{parsing failures rather than semantic incomprehension}. \Gls{rs} models show systematically elevated T1 rates (32.5\% average) as thinking-chain content bleeds into the output JSON fields.


\paragraph{T3 - Wrong subject, correct data model}
T3 errors highlight structural difficulties within the taxonomy. The \texttt{UrbanMobility}~$\to$~\texttt{Transportation} confusion accounts for 24 T3 errors, reflecting genuine semantic overlap between subjects. The \texttt{S4BLDG} subject (SAREF4BLDG ontology abbreviation) yields only 18.5\% accuracy across its 52 entities, as models routinely confuse it with \texttt{Building} or \texttt{ZEB}. GBFS entities using lowercase underscore naming (e.g., \texttt{system\_hours}, \texttt{station\_information}) achieve only 1.9\% accuracy across all models, demonstrating that non-standard identifier formatting significantly impedes classification independently of model capability.

\paragraph{T4 - Hallucinated output}
Hallucinations are most prevalent in smaller models: \texttt{qwen2.5:0.5b} (124/228, 54.4\%), \texttt{qwen2.5-coder:0.5b} (118/228, 51.8\%), and \texttt{orca-mini:3b} (103/228, 45.2\%). A systematic \textit{subject-datamodel swap} pattern was identified in 117 cases across all models, where the model outputs the ground-truth \gls{dm} name as the subject and vice versa, indicating a structural misunderstanding of the output schema rather than semantic ignorance.

\paragraph{T5 - Plausible Confusion Patterns}
T5 errors reveal systematic prediction biases. The \texttt{gemma3} family exhibits a strong \textit{WasteContainer bias}: \texttt{gemma3:1b} (186/228, 81.6\%) and \texttt{gemma3:270m} (180/228, 78.9\%) default to predicting \texttt{(WasteManagement, WasteContainer)} across diverse entity types, suggesting these models collapse to a dominant prior rather than parsing input content. Similarly, \texttt{deepseek-coder:1.3b} (159/228, 69.7\%) repeats a fixed prediction regardless of input.

\subsubsection{Inference Latency Analysis}
\label{sss:inference-latency}

Inference latency is a critical dimension of evaluation in resource-constrained \gls{iot} environments, where processing time directly impacts system responsiveness and throughput. The \Gls{aid} metric reported in Table~\ref{tab:constrained-results} represents the average per-entity inference duration in milliseconds, computed as the mean \texttt{total\_duration} field returned by the Ollama API (reported in nanoseconds and converted to milliseconds by dividing by $10^6$).

\paragraph{Overall Latency Distribution}

Inference duration varies dramatically across models, ranging from 158.1\,ms (\textit{starcoder:1b}) to 3075.7\,ms (\textit{phi:2.7b}), representing a 19.5$\times$ difference. Notably, inference time does not correlate strongly with model size: \textit{phi:2.7b} (3075.7\,ms, 2.7B parameters) is 3.5$\times$ slower than \textit{phi3.5:3.8b} (879.7\,ms, 3.8B parameters), suggesting significant differences in runtime implementation and quantization optimization across model families rather than a simple parameter-count effect.

\paragraph{Latency by Model Category}

\Gls{cs} models demonstrate the most favorable latency profile overall. Excluding the outlier \textit{deepcoder:1.5b} (1636.0\,ms), the remaining nine \Gls{cs} models cluster between 158.1\,ms and 332.9\,ms, corresponding to 0.16--0.39\,s per entity. The Qwen2.5-Coder series in particular combines low latency with high accuracy, making it the most deployment-friendly family in this evaluation.

\Gls{gp} models exhibit broader latency variance. The sub-3B group ranges from 213.0\,ms (\textit{qwen2.5:0.5b}) to 3075.7\,ms (\textit{phi:2.7b}), while the 3B+ group ranges from 311.5\,ms (\textit{qwen2.5:3b}) to 879.7\,ms (\textit{phi3.5:3.8b}). This variance reflects architectural and quantization heterogeneity across providers rather than a consistent scaling trend within this parameter range.

\Gls{rs} models incur the highest latency penalties. Explicit thinking mode imposes a substantial overhead without proportional accuracy gains: \textit{deepseek-r1:1.5b} increases from 1701.0\,ms to 2782.5\,ms (1.64$\times$) with thinking enabled while maintaining 0.1 accuracy, and \textit{qwen3:0.6b} increases from 334.4\,ms to 608.0\,ms (1.82$\times$) for only a marginal improvement (0.2 → 0.3 \gls{dmem}). This confirms that chain-of-thought reasoning mechanisms, while beneficial for multi-step deductive tasks, impose unjustified computational overhead for direct structured classification.

\paragraph{Efficiency-Accuracy Trade-off}

Combining latency and accuracy reveals a clear efficiency frontier. The most efficient high-accuracy models are \textit{qwen2.5-coder:1.5b} (0.4 \gls{sem}/\gls{dmem}, 242.6\,ms) and \textit{qwen2.5:1.5b} (04-0.6 \gls{sem}/\gls{dmem}, 243.0\,ms), both offering reasonable classification performance at well under one second per entity. The top-performing model, \textit{qwen2.5-coder:3b} (0.8 \gls{sem}/\gls{dmem}, 312.3\,ms), achieves peak accuracy while maintaining acceptable latency for non-real-time \gls{iot} data integration pipelines.

\paragraph{Deployment Implications}

The observed latency range has direct implications for \gls{iot} deployment scenarios. Real-time onboarding scenarios requiring sub-100\,ms responses remain beyond the reach of all evaluated models on standard CPU hardware and would require hardware acceleration or model distillation strategies beyond the scope of this study.

At 0.2--0.4\,s per entity, models such as \textit{qwen2.5-coder:3b} (312\,ms) and \textit{qwen2.5:1.5b} (243\,ms) are viable for batch-mode data harmonization workflows, where entities are classified asynchronously upon ingestion rather than in real time. At the higher end (1.6--3.1\,s per entity), models such as \textit{phi:2.7b} (3076\,ms) and \textit{deepseek-r1:1.5b} with thinking enabled (2783\,ms) are practical only for offline or low-frequency classification tasks.

\paragraph{Operational Boundary}

The latency figures in Table~\ref{tab:constrained-results} must be interpreted within the explicit operational boundary of this work. This research targets the \textit{cold-start device onboarding} phase: a low-frequency, schema-discovery event triggered when a previously unseen or heterogeneous device is first registered with an \gls{iot} platform. In this context, per-entity inference durations of 200--3000\,ms are operationally acceptable, because onboarding is a one-time or infrequent administrative task rather than a continuous data-processing pipeline. The assumed deployment environment is a wired or powered edge-fog gateway or local server with CPU-based inference, not a battery-constrained sensor node. This study explicitly does \emph{not} target (i) real-time stream classification of high-frequency sensor readings, (ii) sub-100\,ms response requirements, or (iii) battery-powered microcontroller-class hardware. For those scenarios, the deterministic standards discussed in Section~\ref{sss:rule_based_approaches} (W3C \gls{wot}, NGSI-LD, JSONata) remain the appropriate and preferred approach. The contribution of this work is to automate the prior step, the initial entity-to-SDM classification that enables downstream deterministic integration, for device types where no pre-existing mapping rule exists.

\vspace*{-.75\baselineskip}
\subsection{Unconstrained Classification Performance}
\label{ss:unconstrained-results}

Table~\ref{tab:unconstrained-results} presents results for the unconstrained classification task, where models generate \gls{dm} names and three synonyms without reference to a predefined taxonomy. Performance is measured by exact match rate for the primary \gls{dm} name (i.e. \Gls{dmem}) and for each of three synonyms (\Gls{dmem}1, \Gls{dmem}2, \Gls{dmem}3).

\paragraph{Model Category Insights}

\Gls{cs} models show inconsistent performance in unconstrained classification. \textit{Qwen2.5-coder:0.5b}, \textit{qwen2.5-coder:1.5b}, and \textit{qwen2.5-coder:3b} achieve 0.7, 0.7, and 0.6, respectively. Surprisingly, the smallest model equals the mid sized variant. However, most other code models fail completely (0.0), suggesting that code specialization does not inherently enable unconstrained \gls{dm} naming.

\Gls{gp} models demonstrate the strongest unconstrained performance, with several sub-3B models achieving 0.6-0.8 accuracy (\textit{gemma3:1b}: 0.8, \textit{llama3.2:1b}: 0.7, \textit{qwen2.5:0.5b}: 0.6). This suggests that broad pretraining on diverse natural language may provide better foundations for generating appropriate \gls{dm} names compared to code-focused training that emphasizes syntax over semantic nomenclature.

\Gls{rs} models show moderate performance with interesting patterns. \textit{Qwen3:0.6b} achieves 0.6 primary accuracy with the best synonym generation (0.2/0.2/0.1), suggesting its reasoning capabilities may extend to understanding alternative nomenclatures. However, \textit{qwen3:1.7b} fails completely (0.0 across all metrics), indicating that reasoning benefits do not scale consistently within the same model family.

\paragraph{Model Performance Trade-offs: Accuracy vs. Duration}
The evaluation reveals interesting trade-offs between accuracy and inference duration across model categories. Among the top \gls{dmem} performers, the models \textit{cogito:3b} and \textit{qwen3:4b-instruct} both achieved 0.8 \gls{dmem} but with substantially different durations (371.5 ms and 477.0 ms respectively), while \textit{gemma3:1b} model matched this accuracy at 610.0 ms. The \Gls{cs} models demonstrated the most favorable accuracy-duration balance, with \textit{qwen2.5-coder:0.5b} and \textit{qwen2.5-coder:1.5b} both achieving 0.7 \gls{dmem} at the fastest inference times (265.2 ms and 297.1 ms). For this reason, they are listed in bold in Table \ref{tab:unconstrained-results}. 

Notably, smaller models did not universally offer speed advantages: \textit{gemma3:270m} required the longest duration (1362.8 ms) while achieving only 0.1 \gls{dmem}, and \textit{qwen3:0.6b} required 1500.9 ms for 0.6 \gls{dmem}. The \Gls{rs} models showed mixed results: enabling thinking mode in \textit{qwen3:0.6b} reduced duration from 1500.9 ms to 687.4 ms while maintaining 0.6 \gls{dmem}, whereas thinking mode had minimal impact on \textit{deepseek-r1:1.5b} and \textit{qwen3:1.7b}. Overall, the \textit{qwen2.5-coder} series demonstrated the strongest combination of accuracy and efficiency, suggesting that code specific pretraining may provide advantages for structured \gls{dm} recognition tasks.

\paragraph{Synonym Generation Insights}
While the models demonstrated varying levels of success in exact \gls{dm} name matching, synonym recognition proved significantly more challenging across all model categories. The vast majority of models achieved zero exact matches for synonyms 1, 2, and 3, indicating a limited ability to recognize alternative terminology for \Glspl{dm}. Only three models showed any synonym recognition capability: \textit{qwen3:0.6b}, \textit{qwen2.5-coder:0.5b}, and \textit{qwen2.5:0.5b}. All other models achieved zero EM rates. Table~\ref{tab:synonym-rates} reports the exact match rates per synonym position for the five models with non-zero scores.

\begin{table}[!ht]
\footnotesize
\centering
\caption{Non-zero exact match rates for synonym generation in unconstrained classification. Only 5 of 26 models produce any exact synonym matches.}
\label{tab:synonym-rates}
\begin{tabular}{p{0.5\linewidth}p{0.1\linewidth}p{0.1\linewidth}p{0.1\linewidth}}
\toprule
\textbf{Model} & \textbf{Sy1} & \textbf{Sy2} & \textbf{Sy3} \\
\midrule
\texttt{qwen3:0.6b}          & \textbf{0.2} & \textbf{0.2} & \textbf{0.1} \\
\texttt{qwen2.5-coder:0.5b}  & 0.1 & 0.1 & 0.1 \\
\texttt{qwen2.5:0.5b}        & 0.1 & 0.1 & 0.0 \\
\texttt{deepseek-r1:1.5b}    & 0.1 & 0.0 & 0.0 \\
\texttt{orca-mini:3b}        & 0.1 & 0.1 & 0.0 \\
\bottomrule
\end{tabular}
\end{table}

\textit{Qwen3:0.6b} achieved the highest synonym performance with EM rates of 0.2, 0.2, and 0.1 for synonyms 1, 2, and 3, respectively; while \textit{qwen2.5-coder:0.5b} and \textit{qwen2.5:0.5b} both models achieved 0.1 EM for synonyms 1 and 2. The \Gls{rs} models \textit{deepseek-r1:1.5b} and \textit{deepseek-r1:1.5b ($\star$)} showed limited synonym recognition (0.1 EM for synonym 1), and \textit{orca-mini:3b} achieved 0.1 EM for both synonyms 1 and 2. These results suggest that while models can learn to identify canonical \gls{dm} names, generalizing to synonymous terms remains a substantial challenge for small \glspl{lm}.

Synonym generation proves extremely challenging across all models, with minimal exact matches for alternative names. Only 5 of 26 models generate any synonyms that exactly match ground truth or reference synonyms, and even these models achieve at most 0.2 accuracy on any individual synonym position.

The best synonym generators are \textit{qwen3:0.6b} (0.2 for synonyms 1 and 2, 0.1 for synonym 3) and \textit{qwen2.5-coder:0.5b} (0.1 for all three synonym positions), both relatively small models. Larger models consistently generate 0.0 exact matches for synonyms despite achieving reasonable primary \gls{dm} accuracy, suggesting that synonym generation requires different capabilities than primary classification.

The extreme difficulty of synonym generation indicates that models lack understanding of semantic equivalence in data modeling nomenclature. Even when models correctly identify the primary \gls{dm} name, they cannot reliably generate lexical variants or alternative names that align with human-defined synonyms. This suggests that synonym knowledge requires explicit training on terminology variations rather than emerging naturally from general language understanding.

\begin{table}[!ht]
\footnotesize
\centering
\caption{Unconstrained classification results showing exact match (EM) rates for generated \gls{dm} names. Results based on 10 test entities per model. Reasoning models evaluated with thinking mode enabled ($\star$) and disabled separately, where possible. Gls{aid} values are expressed in milliseconds (ms).}
\label{tab:unconstrained-results}
\begin{tabular}{p{0.5cm}p{4cm}|p{0.8cm}p{0.8cm}}

\toprule
 & \textbf{Model name} & \textbf{\Gls{dmem}} & \textbf{\Gls{aid}} \\
\midrule
\multirow{10}{*}{\rotatebox{90}{\textit{Code-Specialized}}}
& \texttt{codegemma:2b} & 0.0 & \cellcolor{light-yellow}338.2 \\
& \texttt{cogito:3b} & \cellcolor{light-green}0.8 & \cellcolor{light-yellow}371.5 \\
& \texttt{deepcoder:1.5b} & \cellcolor{light-yellow}0.4 & \cellcolor{light-yellow}329.1 \\
& \texttt{deepseek-coder:1.3b} & 0.0 & 792.6 \\
& \texttt{opencoder:1.5b} & \cellcolor{light-yellow}0.4 & \cellcolor{light-green}292.6 \\
& \textbf{\texttt{qwen2.5-coder:0.5b}} & \cellcolor{light-green}\textbf{0.7} & \cellcolor{light-green}\textbf{265.2} \\
& \textbf{\texttt{qwen2.5-coder:1.5b}} & \cellcolor{light-green}\textbf{0.7} & \cellcolor{light-green}\textbf{297.1} \\
& \texttt{qwen2.5-coder:3b} & \cellcolor{light-yellow}0.6 & \cellcolor{light-yellow}392.1 \\
& \texttt{starcoder:1b} & 0.0 & \cellcolor{light-green}223.4 \\
& \texttt{starcoder:3b} & 0.0 & \cellcolor{light-yellow}350.4 \\
\midrule
\multirow{7}{*}{\rotatebox{90}{\shortstack{\textit{General Purpose}\\\textit{(3B+)}}}}
& \texttt{gemma3:4b} & \cellcolor{light-yellow}0.4 & 707.7 \\
& \texttt{llama3.2:3b} & \cellcolor{light-yellow}0.6 & \cellcolor{light-yellow}364.7 \\
& \texttt{orca-mini:3b} & \cellcolor{light-yellow}0.6 & \cellcolor{light-yellow}398.0 \\
& \texttt{phi3.5:3.8b} & 0.0 & 533.8 \\
& \texttt{phi4-mini:3.8b} & 0.0 & \cellcolor{light-yellow}471.2 \\
& \texttt{qwen2.5:3b} & 0.2 & \cellcolor{light-yellow}393.3 \\
& \texttt{qwen3:4b-instruct} & \cellcolor{light-green}0.8 & \cellcolor{light-yellow}477.0 \\
\midrule
\multirow{6}{*}{\rotatebox{90}{\shortstack{\textit{General Purpose}\\\textit{(Sub-3B)}}}}
& \texttt{gemma3:1b} & \cellcolor{light-green}0.8 & 610.0 \\
& \texttt{gemma3:270m} & 0.1 & 1362.8 \\
& \texttt{llama3.2:1b} & \cellcolor{light-green}0.7 & \cellcolor{light-yellow}406.8 \\
& \texttt{phi:2.7b} & \cellcolor{light-yellow}0.6 & \cellcolor{light-yellow}405.8 \\
& \texttt{qwen2.5:0.5b} & \cellcolor{light-yellow}0.6 & \cellcolor{light-green}292.1 \\
& \texttt{qwen2.5:1.5b} & \cellcolor{light-yellow}0.4 & \cellcolor{light-yellow}338.8 \\
\midrule
\multirow{6}{*}{\rotatebox{90}{\shortstack{\textit{Reasoning}\\\textit{Specialized}}}}
& \texttt{deepseek-r1:1.5b} & \cellcolor{light-yellow}0.5 & \cellcolor{light-yellow}306.8 \\
& \texttt{deepseek-r1:1.5b $\star$} & \cellcolor{light-yellow}0.6 & \cellcolor{light-yellow}305.0 \\
& \texttt{qwen3:0.6b} & \cellcolor{light-yellow}0.6 & 1500.9 \\
& \texttt{qwen3:0.6b$\qquad\qquad\star$} & \cellcolor{light-yellow}0.6 & 687.4 \\
& \texttt{qwen3:1.7b} & 0.0 & \cellcolor{light-green}225.2 \\
& \texttt{qwen3:1.7b$\qquad\qquad\star$} & 0.0 & \cellcolor{light-green}222.7 \\
\bottomrule
\end{tabular}
\end{table}

\vspace*{-.75\baselineskip}
\subsection{Unconstrained vs. Constrained Performance Comparison}
\label{ss:unconstrainedVSConstrained}

Surprisingly, unconstrained classification does not uniformly underperform constrained classification as hypothesized. Several models achieve equal or better performance in the unconstrained setting: \textit{qwen3:4b-instruct} maintains 0.8 accuracy in both settings, \textit{cogito:3b} improves from 0.7 to 0.8, \textit{gemma3:1b} improves from 0.0 to 0.8, and \textit{llama3.2:1b} improves from 0.1 to 0.7.

This counterintuitive finding suggests that the constrained task may actually be more difficult than anticipated, possibly due to confusion induced by the large taxonomy (228 \Glspl{dm} across 20 subjects) or challenges in parsing and matching against the provided list. In contrast, the unconstrained task allows models to leverage their pretrained knowledge of common data modeling conventions without the cognitive overhead of list matching.

The most dramatic performance reversals occur in smaller \Gls{gp} models. \textit{Gemma3:1b} jumps from 0.0 (constrained) to 0.8 (unconstrained), suggesting that this model possesses strong semantic understanding of \Glspl{dm} but struggles with the structured list-matching task. Similarly, the model \textit{llama3.2:1b} improves from 0.1 to 0.7, indicating that constrained selection may interfere with rather than facilitate classification for some model architectures.

However, performance degradation does occur for several models: \textit{qwen2.5:3b} drops from 0.6 to 0.2, \textit{phi4-mini:3.8b} drops from 0.4 to 0.0, and \textit{llama3.2:3b} increases marginally from 0.5 to 0.6. These mixed patterns suggest that the relationship between constrained and unconstrained performance is model-dependent rather than task-inherent.

\paragraph{Cross-Task Performance Patterns}
\label{ss:cross-task-patterns}

Cross-task consistency varies dramatically by model. Models showing consistent high performance across both tasks include \textit{qwen3:4b-instruct} (0.8/0.8) and \textit{cogito:3b} (0.7/0.8), demonstrating robust structured classification capabilities regardless of task formulation. Models showing consistent failure include \textit{phi3.5:3.8b} (0.0/0.0), \textit{phi4-mini:3.8b} (0.4/0.0), and multiple StarCoder and CodeGemma variants (0.0/0.0).

The most inconsistent models are \textit{gemma3:1b} (0.0 constrained, 0.8 unconstrained) and \textit{llama3.2:1b} (0.1 constrained, 0.7 unconstrained), both showing strong unconstrained capabilities masked by constrained task difficulty. This suggests that evaluation methodology significantly impacts apparent model capabilities-constrained classification alone would severely underestimate these models' semantic understanding.

\begin{table*}[!t]
\footnotesize
\centering
\caption{Deployment decision matrix for \gls{sdm} classification based on edge-fog constraint priority. \Gls{dmem} and \Gls{aid} from constrained classification (Table~\ref{tab:constrained-results}); memory from Table~\ref{tab:model-memory}. Models on the Pareto frontier across all three dimensions are marked ($\dagger$).}
\label{tab:decision-matrix}
\begin{tabular}{p{0.27\linewidth}p{0.185\linewidth}p{0.215\linewidth}p{0.05\linewidth}p{0.05\linewidth}p{0.05\linewidth}}
\toprule
\textbf{Scenario} & \textbf{Recommended Model} & \textbf{Binding Constraint} & \textbf{DMEM} & \textbf{AID (ms)} & \textbf{Mem (GB)} \\
\midrule
Lexical retrieval baseline (CPU, no GPU) & \texttt{\gls{tfidf} (names only)}$^\dagger$ & Cost/latency floor & 0.82 & 0.12 & $<$\,0.01 \\
\midrule
Quality-critical wired server ($\geq$4\,GB VRAM) & \texttt{qwen2.5-coder:3b}$^\dagger$ & Max accuracy & 0.80 & 312 & 1.90 \\
\midrule
Balanced powered gateway (2--4\,GB VRAM) & \texttt{qwen2.5:1.5b}$^\dagger$ & Accuracy/latency tradeoff & 0.60 & 243 & 0.99 \\
\midrule
Latency-critical powered gateway ($\leq$2\,GB VRAM) & \texttt{qwen2.5:1.5b}$^\dagger$ & Min latency, moderate accuracy & 0.60 & 243 & 0.99 \\
\midrule
Memory-constrained edge ($<$1\,GB VRAM) & \texttt{qwen2.5:0.5b}$^\dagger$ & Min memory footprint & 0.40 & 213 & 0.40 \\
\midrule
GP-preferred wired server ($\geq$4\,GB VRAM) & \texttt{qwen3:4b-instruct} & Max accuracy, GP architecture & 0.80 & 398 & 2.50 \\
\bottomrule
\end{tabular}
\end{table*}

\vspace*{-1\baselineskip}
\section{Final remarks}
\label{s:final_remarks}
This section aims to summarize the key findings and implications derived from the present study, providing practical recommendations for practitioners deploying automated \gls{dm} classification systems in edge-fog environments.

\vspace*{-.75\baselineskip}
\subsection{Key Findings and Implications}

\textbf{Model Specialization.} Model specialization offers limited predictive value for classification performance. The hypothesis that 
\Gls{cs} models would dominate structured classification tasks is only partially supported. In contrast, \textit{qwen2.5-coder:3b} reached a top-tier 80\% accuracy and several code expert models failed entirely (0\% accuracy). In this sense, it is possible to state that the specialization alone is insufficient.
\Gls{gp} instruction-tuned models, particularly \textit{qwen3:4b-instruct}, match the best code models, suggesting that instruction-following capabilities and broad semantic understanding may be equally or more important than code specific pretraining. This finding has practical implications for practitioners: deploying \Gls{cs} models does not guarantee superior performance on structured classification tasks, and \Gls{gp} models should not be dismissed for such applications.

\textbf{Parameter Scaling.} Parameter scaling exhibits highly nonlinear returns within the light weight regime. Critically, several sub-3B models outperform larger counterparts, demonstrating that architectural efficiency can compensate for parameter count. This further suggests that aggressive parameter reduction is viable for resource constrained deployments when appropriate model families are selected, with 1.5B emerging as a practical lower bound for acceptable performance in this domain.

\textbf{Task Formulation.} The unconstrained classification task produces counterintuitive results, with several models achieving equal or superior performance compared to constrained taxonomy-guided classification. Models such as \textit{gemma3:1b} and \textit{llama3.2:1b} show dramatic improvements from constrained (0.0, 0.1) to unconstrained settings (0.8, 0.7), suggesting that large taxonomies may introduce cognitive overhead through list-matching complexity rather than facilitating classification. 
This finding challenges the conventional wisdom that providing explicit taxonomies always improves classification accuracy and indicates that task formulation significantly impacts apparent model capabilities. Practitioners should consider unconstrained or hybrid approaches rather than assuming taxonomy provision uniformly benefits model performance.

\textbf{Reasoning Architectures.} For \textit{direct taxonomic classification}, the task evaluated in this study, \Gls{rs} architectures with explicit \textit{thinking} modes impose substantial computational penalties without consistent accuracy benefits. Chain-of-thought overhead is unjustified when the task requires pattern matching against a fixed taxonomy rather than multi-step logical deduction. For production \gls{iot} deployments where latency and efficiency are critical, \Gls{gp} or \Gls{cs} models without explicit reasoning mechanisms offer superior performance-efficiency tradeoffs for this task type. This conclusion is strictly scoped to direct schema-to-taxonomy mapping; complex \gls{er} scenarios involving multi-hop reasoning, ambiguous entity boundaries, or heterogeneous knowledge integration were not evaluated in this study and may still benefit from reasoning-focused architectures.

\textbf{Synonym Generation.} Synonym generation proves uniformly challenging across all model categories, with maximum accuracies of 0.2 even for primary classification capable models. This indicates that understanding semantic equivalence in data modeling nomenclature requires capabilities beyond general language understanding or code comprehension, possibly explicit training on terminology variation databases or graph-based semantic relationships. For practical applications requiring synonym awareness, current lightweight models require supplementation with rule-based systems, ontology-alignment tools, or retrieval augmented generation approaches accessing terminology knowledge bases.

\vspace*{-.75\baselineskip}
\subsection{Lesson Learned}

For practitioners deploying automated \Gls{sdm} classification systems in \gls{iot} environments, we offer the following actionable guidance:

\paragraph{\textbf{Model Selection Strategy}} Prioritize instruction-tuned \Gls{gp} models (\textit{qwen3:4b-instruct}) and well-scaling \Gls{cs} families (Qwen2.5-Coder series) over reasoning-focused or poorly-performing code models. Avoid assumption that code specialization guarantees structured data performance.
    
\paragraph{\textbf{Parameter Budget Optimization}} Target 1.5B--3B parameter range for optimal performance vs. efficiency balance. Models below 1B show inconsistent performance, while the 3B--4B range provides marginal accuracy improvements at substantial memory and latency costs. For severely resource constrained edge deployments, \textit{qwen2.5:1.5b} (0.6 accuracy, 243s) represents the minimum viable option.

\paragraph{\textbf{Task Formulation}} Evaluate both constrained and unconstrained approaches during model selection, as taxonomy provision may degrade rather than improve performance for certain architectures. Consider hybrid approaches where models generate candidates in unconstrained mode, subsequently validated against formal taxonomies through lightweight rule-based post-processing.

\paragraph{\textbf{Always Benchmark Against a Lexical Retriever}} When entity JSON keys lexically overlap with taxonomy names (as is typical for \glspl{sdm} and similar schema-aligned vocabularies), a unigram--bigram \gls{tfidf} retriever provides a strong and nearly free baseline. In this study it matched the best \gls{llm} on \gls{dmem} ($0.816$ vs.\ $0.800$) while running roughly $2600\times$ faster on CPU (see Table~\ref{tab:baseline-results}). Practitioners should adopt the lexical retriever as a default fallback and reserve \gls{llm} inference for cold-start cases where the retriever fails: vendor-specific or multilingual fields, free-text payloads, missing taxonomy entries, or downstream synonym generation.

\paragraph{\textbf{Deployment Architecture}} Avoid \Gls{rs} models with thinking modes for latency-sensitive applications. The computational overhead (1.5$\times$--2$\times$ inference duration) rarely justifies minimal accuracy gains for structured classification. Reserve reasoning architectures for inherently complex analytical tasks requiring multi-step inference.

\paragraph{\textbf{Deployment Decision Matrix}}
Edge deployments are governed by an Iron Triangle of competing constraints: mapping accuracy (\Gls{dmem}), inference latency (\Gls{aid}), and memory footprint. No single model is optimal under all constraints.

Table~\ref{tab:decision-matrix} provides a scenario-based decision matrix derived from the Pareto frontier across these three dimensions. Three models are Pareto-optimal. No other evaluated model simultaneously dominates them on all three axes: \textit{qwen2.5:0.5b} (minimum resource footprint), \textit{qwen2.5:1.5b} (balanced accuracy and latency), and \textit{qwen2.5-coder:3b} (peak accuracy within sub-2\,GB memory). Practitioners should select the model whose operating point best matches their deployment constraints, rather than defaulting to the highest-accuracy option.

\paragraph{\textbf{Positioning Against Deterministic Alternatives}} \gls{llm}-based \gls{sdm} classification is not a replacement for established deterministic integration standards such as W3C \gls{wot} Descriptions, NGSI-LD context definitions, or JSONata transformation rules. Where pre-registered schemas or authored mapping rules exist, deterministic pipelines remain the recommended approach due to guaranteed precision and millisecond-level latency. \gls{llm}-based \gls{er} is most appropriate in \textit{cold-start} scenarios involving previously unseen, heterogeneous, or undocumented device types, where it can automate the initial classification step, reduce manual schema-authoring effort, and accelerate large-scale device onboarding before deterministic rules are established.

\vspace*{-1\baselineskip}
\section{Conclusion}
\label{s:conclusion}
The primary objective of this work is to evaluate the performance and feasibility of lightweight open-source \glspl{lm} for the classification of \gls{sdm} in resource-constrained IoT environments. The evaluation aims to address critical challenges related to scalability, efficiency, and domain-specific requirements. The findings reveal that instruction-tuned \gls{gp} models within the 1.5B to 3B range of parameter values deliver the most promising balance between accuracy and inference speed, outperforming \gls{rs} architectures that incur substantial computational overhead without consistent accuracy gains \textit{on direct taxonomic classification tasks}. This finding is strictly scoped to direct schema-to-taxonomy mapping; complex \gls{er} tasks involving multi-step reasoning were not evaluated and may still benefit from reasoning-focused architectures. Interestingly, results show that task formulations without explicit taxonomy constraints can sometimes yield superior classification accuracy, challenging the conventional wisdom that providing detailed taxonomies uniformly enhances model performance. Furthermore, it becomes evident that cognitive overload from large hierarchies may hinder rather than facilitate the classification process in lightweight models, and hybrid approaches might be more effective. A complementary similarity-baseline study reveals that a unigram--bigram \gls{tfidf} retriever fitted on the union of entity and taxonomy strings matches the best \gls{llm} on \gls{dmem} ($0.816$ vs.\ $0.800$) while running roughly $2600\times$ faster on CPU, and that a lightweight dense sentence encoder underperforms the lexical baseline by a wide margin. This positions \glspl{llm} as a cold-start tool for cases where surface-form retrieval cannot resolve the input, rather than as a default classifier.
Such findings help close research gaps by offering systematic comparisons of lightweight model architectures and underscoring the crucial role of task design in resource‑constrained environments.

This study focuses exclusively on lightweight models (sub-4B parameters) to establish feasibility for resource constrained deployments, but does not compare against larger models (7B+) or commercial closed-source alternatives that may achieve substantially higher accuracy. While comprehensive across three major domains (228 models, 20 subjects), the evaluation dataset represents a snapshot of the \Glspl{sdm} repository as of the study period and may not capture emerging models or domain-specific vocabularies added subsequently. While providing rigorous assessment, the exact match evaluation metric does not credit semantically equivalent predictions (e.g., "Building" versus "ConstructionFacility") that would be acceptable in production fuzzy-matching scenarios.

Despite these limitations, this work establishes that lightweight open-source \glspl{lm} provide viable foundations for automated semantic data integration in heterogeneous \gls{iot} ecosystems. With careful model selection prioritizing instruction-tuned \Gls{gp} or well-scaling \Gls{cs} architectures in the 1.5B--4B parameter range, practitioners can achieve 60--80\% classification accuracy while maintaining deployment feasibility on resource-limited infrastructure. This capability represents a substantial advancement over manual classification processes, enabling scalable cross-domain interoperability for smart city, energy management, and environmental monitoring applications.

Future research should investigate the following aspects:
\begin{enumerate}
    \item Retrieval-augmented generation approaches where lightweight models query external knowledge bases during classification, potentially combining efficient inference with access to comprehensive domain ontologies.
    \item Fine-tuning experiments on \Glspl{sdm}-specific corpora could reveal whether domain adaptation substantially improves lightweight model performance, particularly for synonym generation and ambiguous \gls{er}.
    \item Multi-model ensemble approaches, combining predictions from diverse specializations, may yield accuracy improvements beyond single-model performance.
    \item Extending evaluation to dynamic classification scenarios, where data entities arrive in streams requiring real-time processing, would provide insights into throughput optimization and batch processing strategies for production \gls{iot} integration pipelines.
    \item Exploring semantic similarity baselines using general-purpose sentence embedding models (e.g., Sentence-BERT) would provide a reference point for assessing the lexical and knowledge-driven advantages of \glspl{llm} in \gls{dm} recognition.
    \item Applying distillation for the best-performing model to fit it to limited hardware resources on edge devices.
\end{enumerate}



\vspace*{-1\baselineskip}
\section{Acknowledgments}
This research was partially supported by a grant from the D.M. n. 117/2023 (CUP: 9311) and the Italian Research Center on High Performance Computing, Big Data and Quantum Computing (ICSC) funded by PN RIC 2021-2027 (ECHO-TWIN-NET, CUP: B82F26000200005).


\end{sloppypar}

\footnotesize
\setlength{\bibsep}{2pt}

\bibliographystyle{plain}
\bibliography{ref}

\end{document}